\documentclass[lettersize,journal]{IEEEtran}
\usepackage{amsmath,amsfonts}
\usepackage{algorithmic}
\usepackage{algorithm}
\usepackage{array}
\usepackage[caption=false,font=normalsize,labelfont=sf,textfont=sf]{subfig}
\usepackage{textcomp}
\usepackage{stfloats}
\usepackage{url}
\usepackage{verbatim}
\usepackage{graphicx}
\usepackage{cite}

\usepackage{multirow}
\usepackage{multicol}
\usepackage{booktabs}
\usepackage{threeparttable}
\usepackage[table]{xcolor}
\usepackage{hyperref}
\hypersetup{
    colorlinks=true,
    linkcolor=blue,
    urlcolor=blue,
    citecolor=blue}

\begin{document}

\title{Downstream Task Inspired Underwater Image Enhancement: A Perception-Aware Study from Dataset Construction to Network Design}

\author{Bosen Lin, 
Feng Gao, \emph{Member}, \emph{IEEE},
Yanwei Yu, \emph{Member}, \emph{IEEE},
Junyu Dong, \emph{Member}, \emph{IEEE}, \\
Qian Du, \emph{Fellow}, \emph{IEEE}
\thanks{This work was supported in part by the National Science and Technology Major Project under Grant 2022ZD0117201, in part by the Natural Science Foundation of Shandong Province under Grant ZR2024MF020. (\textit{Corresponding author: Feng Gao})

Bosen Lin, Feng Gao, Yanwei Yu, and Junyu Dong are with the State Key Laboratory of Physical Oceanography, Ocean University of China, Qingdao 266100, China. 

Qian Du is with the Department of Electrical and Computer Engineering, Mississippi State University, Starkville, MS 39762 USA.

Digital Object Identifier 10.1109/TIP.2025.XXXXXXX}}

\markboth{IEEE TRANSACTIONS ON IMAGE PROCESSING}{}

\maketitle

\begin{abstract}
In real underwater environments, downstream image recognition tasks such as semantic segmentation and object detection often face challenges posed by problems like blurring and color inconsistencies. Underwater image enhancement (UIE) has emerged as a promising preprocessing approach, aiming to improve the recognizability of targets in underwater images. However, most existing UIE methods mainly focus on enhancing images for human visual perception, frequently failing to reconstruct high-frequency details that are critical for task-specific recognition. To address this issue, we propose a Downstream Task-Inspired Underwater Image Enhancement (DTI-UIE) framework, which leverages human visual perception model to enhance images effectively for underwater vision tasks. Specifically, we design an efficient two-branch network with task-aware attention module for feature mixing. The network benefits from a multi-stage training framework and a task-driven perceptual loss. Additionally, inspired by human perception, we automatically construct a Task-Inspired UIE Dataset (TI-UIED) using various task-specific networks. Experimental results demonstrate that DTI-UIE significantly improves task performance by generating preprocessed images that are beneficial for downstream tasks such as semantic segmentation, object detection, and instance segmentation. The code will be made publicly available at \url{https://github.com/oucailab/DTIUIE}.
\end{abstract}

\begin{IEEEkeywords}
Underwater Image Enhancement, Image Recognition, Human Vision Perception, Underwater Dataset.
\end{IEEEkeywords}

\section{Introduction}

Underwater vision is essential for underwater robots to perceive their environment and acquire target information \cite{hongUSOD10KNewBenchmark2025}\cite{zhaoCompositedFishNetFish2021}. However, light absorption, reflection, and scattering in water reduce image clarity and cause distortion \cite{liUnderwaterImageEnhancement2020} \cite{liuRealWorldUnderwaterEnhancement2020} \cite{muGeneralizedPhysicalknowledgeguidedDynamic2023}\cite{quanEnhancingUnderwaterImages2024b}. Consequently, underwater images exhibit visual characteristics that differ significantly from natural images, posing substantial challenges for downstream vision tasks. This problem becomes particularly acute when image recognition methods developed for natural images are deployed on underwater platforms \cite{waszakSemanticSegmentationUnderwater2022}.

\begin{figure}[h]
  \centering
  \includegraphics[width=\linewidth]{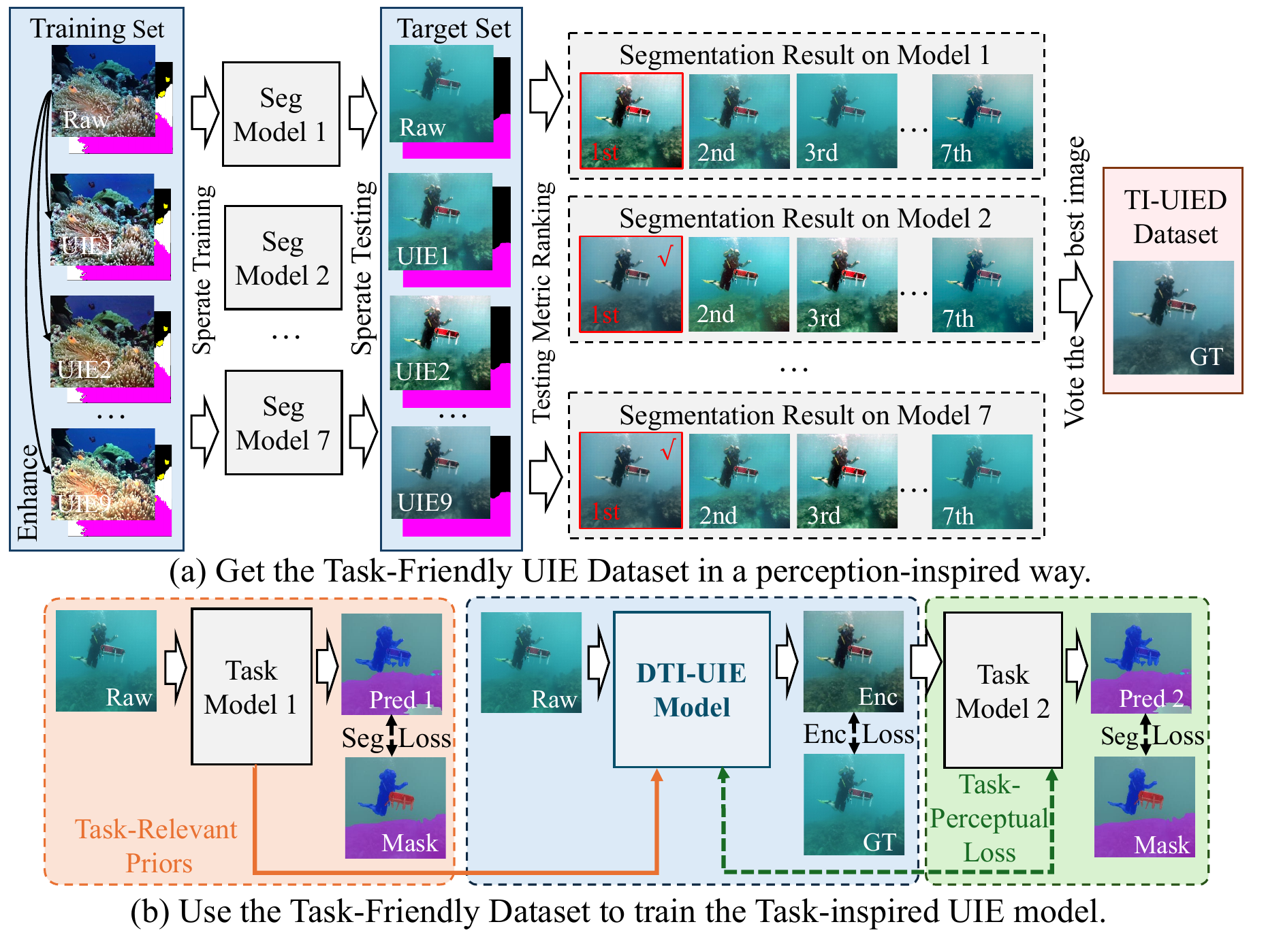}
  \caption{Framework of the proposed DTI-UIE. (a) A task-inspired UIE dataset is constructed automatically by task networks. (b) A UIE network providing advantages for downstream tasks is achieved with task-relevant priors and task-driven perceptual loss.}
  \label{Figure:overall}
\end{figure}

It is common knowledge that improving the visual quality of images can improve the performance of downstream tasks \cite{wuFlexibleSemanticGuided2024}\cite{kimImageSuperResolutionImage2024a}. Consequently, Underwater Image Enhancement (UIE) methods have been proposed as a preprocessing step for recognition tasks \cite{yuTaskFriendlyUnderwaterImage2024}, including underwater object detection \cite{zhouAMSPUODWhenVortex2024} \cite{fuLearningHeavilyDegradedPrior2023b} and semantic segmentation \cite{islamSemanticSegmentationUnderwater2020}. These methods leverage traditional image processing algorithms or deep neural networks to enhance underwater images, aiming to provide improved input for subsequent tasks. However, recent research indicates that state-of-the-art UIE algorithms do not consistently translate to improved performance in downstream tasks \cite{wangUnderwaterImageEnhancement2023}. 

\begin{figure}[t]
  \centering
  \includegraphics[width=3.0in]{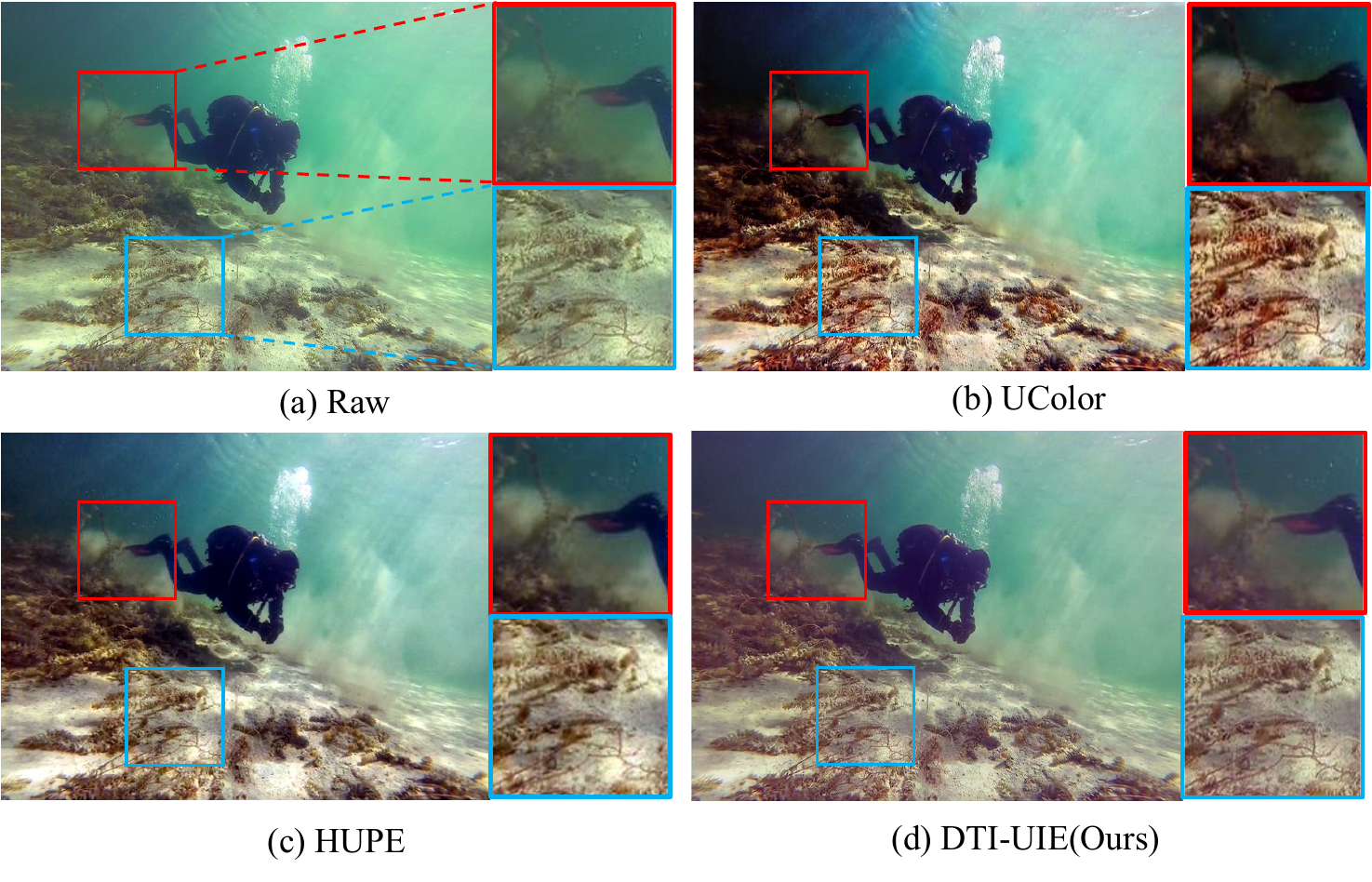}
  \caption{Underwater images enhanced using different UIE methods. UColor \cite{liUnderwaterImageEnhancement2021} and HUPE\cite{zhangHUPEHeuristicUnderwater2025} blurred the outlines of the diver and background, and introduced additional edge and texture information to the foreground seabed. The proposed DTI-UIE introduced more accurate foreground-background differentiation and reduced texture and edge noise, which is beneficial for downstream tasks.}\label{Figure:IntroImg}
\end{figure}

It is a non-trivial task to develop an effective UIE framework that systematically enhances downstream task's accuracy, due to the following challenges: \textbf{(1) \textit{Lack of dedicated underwater image datasets tailored for downstream tasks.}}  While many existing datasets focus on visual quality improvement, they are not designed with the requirements of object detection and segmentation. As a result, enhanced images may exhibit better visual appearance but fail to improve, or may even degrade, the performance of subsequent high-level tasks. This gap highlights the urgent need for benchmark datasets that explicitly consider the interaction between enhancement and task performance in underwater scenarios. \textbf{(2) \textit{Misalignment between enhancement objectives and downstream task requirements.}} Most existing UIE frameworks are primarily optimized for human visual perception, focusing on improving image aesthetics such as contrast, color balance, and visibility. However, these methods may introduce enhancement-induced artifacts, such as amplifying back-scattered noise in low-visibility background areas, or producing over-sharpened boundaries and spurious edges. These phenomena distort critical textures and edge structures, blurring task-relevant high-frequency information \cite{yuTaskFriendlyUnderwaterImage2024}\cite{wangUnderwaterImageEnhancement2023}\cite{saleemUnderstandingInfluenceImage2025}. Visualizations of some enhanced images geared towards human perception are shown in Figure \ref{Figure:IntroImg}. Although the colors and contrasts of these images are more realistic, different levels of blurring are present in key areas that distinguish the target from the background. 

To solve the first challenge, we collect and construct the \textbf{T}ask-\textbf{I}nspired \textbf{UIE} \textbf{D}ataset (\textbf{TI-UIED}), which is specifically tailored for downstream vision tasks rather than subjective human preferences. Instead of relying on manual annotations or perceptual voting \cite{zhangHUPEHeuristicUnderwater2025} \cite{qiSGUIENetSemanticAttention2022}, we adopt an automatic, task-oriented construction pipeline, as illustrated in Fig. \ref{Figure:overall}(a). In particular, for each underwater scene, multiple enhanced versions are generated using a range of UIE methods. Their effectiveness is then evaluated across diverse semantic segmentation networks. The enhanced image that achieves the most consistent improvement in segmentation accuracy is then selected as the task-driven ground truth. This approach ensures that the TI-UIED better preserves task-relevant features and provides meaningful supervision for real-world downstream applications. 

To bridge the gap between enhancement goals and the needs of downstream tasks, we propose a \textbf{D}ownstream \textbf{T}ask-\textbf{I}nspired \textbf{U}nderwater \textbf{I}mage \textbf{E}nhancement (\textbf{DTI-UIE}) framework, as illustrated in Fig. \ref{Figure:overall}(b).
Inspired by the advanced processing mechanisms of human vision \cite{liUnderwaterImageEnhancement2020} \cite{pengUShapeTransformerUnderwater2023}, we design a dual-branch enhancement network. The network consists of a multi-scale Conv-attention Transformer encoder-decoder \cite{xuCoScaleConvAttentionalImage2021} for task-related feature extraction,  with a resolution-preserving subnet designed to recover fine-grained edges and textures. Additionally, inspired by the brain's ability to use past experiences for object recognition, we introduce the Task-Aware Conv-attention Transformer Block (TA-CTB), a novel module that injects task-specific priors into the enhancement process. By enabling cross-model interaction between heterogeneous task representations, TA-CTB strengthens the preservation of features that are critical for recognition. 

Furthermore, we design a multi-stage training framework that emulates integration and adaptive response mechanisms of human perception. In the first stage, a task network is trained to generate feature priors that encapsulate target-relevant information. These priors guide the enhancement process in the second stage, where we optimize the enhancement network using both pixel-level loss and Task-Driven Perceptual (TDP) loss. The TDP loss measures feature discrepancies using a task network, as well as compels the enhancement network to restore details in the feature space that are essential for recognition accuracy. Finally, in the third stage, we dynamically update the task loss network with mixed-image inputs to prevent shortcut learning and improve generalization. Across all stages, the feature prior network, enhancement network, and task loss network are alternately optimized, creating a tightly coupled learning cycle. 

In summary, the contributions of this paper are summarized as follows:

\begin{itemize}

\item We construct the TI-UIED, the first UIE dataset explicitly tailored for downstream vision tasks. Unlike existing datasets focused on human visual perception, TI-UIED is automatically generated by selecting enhanced images that yield superior and consistent performance across multiple segmentation networks, ensuring that the reference reflects task-oriented requirements. 

\item We propose the DTI-UIE framework to enhance the performance of downstream tasks. DTI-UIE employs a dual-branch architecture: a conv-attention Transformer encoder–decoder to extract and reconstruct semantic features, and a resolution-preserving subnet to refine edges and textures. Furthermore, we introduce a novel TA-CTB to integrate task-specific priors from task networks, thereby enabling effective cross-modal feature interaction.

\item To achieve stronger alignment between enhancement objectives and downstream task requirements, we introduce a TDP loss along with a three-stage training strategy. This framework tightly couples feature learning with task-level supervision. Extensive experiments on semantic segmentation, object detection, and instance segmentation demonstrate that our method consistently outperforms existing UIE approaches.

\end{itemize}

The rest of the paper is organized as follows. The related work about both UIE and UIE datasets are  reviewed in Section \ref{Section:2}. The construction and characteristics of the proposed TI-UIED is detailed in \ref{Section:3}. The detailed structure of the proposed DTI-UIE network is introduced in Section \ref{Section:4}. The experimental setup and results are reported in Section \ref{Section:5}. Finally we give the discussion in Section \ref{Section:6} and a conclusion in Section \ref{Section:7}.

\section{Related Work}\label{Section:2}

\subsection{UIE inspired by human visual perception}

Underwater images often suffer from poor quality due to water’s absorption and scattering effects, resulting in issues like color cast, low contrast, blurriness, uneven brightness, and underexposure. To enhance visibility, traditional UIE methods adjust pixels individually. Zhang et al. \cite{zhangUnderwaterImageEnhancement2024} proposed the Weighted Wavelet Visual Perception Fusion (WWPF) method, which fuses high- and low-frequency components at multiple scales. Jha et al. \cite{jhaCBLAColorBalancedLocally2024} introduced the color-balanced locally adjustable (CBLA) method, improving color balance in both RGB and CIELab spaces.\textit{ However, these methods often introduce unnatural colors and corrupted edge information due to their limited adaptability to complex underwater environments.}

Recent advances in deep learning have enabled perception-inspired UIE. For raw underwater images, pseudo ground-truths are initially generated by inviting human volunteers manually vote on the optimal enhancement \cite{liUnderwaterImageEnhancement2020}. The transformation from the raw image to the ground-truth is then learned through neural networks \cite{pengUShapeTransformerUnderwater2023}. Qi et al. \cite{qiDeepColorCorrectedMultiscale2024} proposed the deep color-corrected multiscale Retinex network (CCMSR-Net) based on the Retinex physical imaging model. Huang et al. \cite{huangContrastiveSemisupervisedLearning2023} introduced the Semi-supervised Underwater Image Restoration (Semi-UIR) method, incorporating the mean-teacher approach to leverage unlabeled data in UIE networks' training. Guo et al. \cite{guoUnderwaterRankerLearn2022} employed a pretrained image quality assessment method as additional supervision in the training stage of UIE network. Zhao et al. \cite{zhaoWaveletbasedFourierInformation2024} achieved detailed UIE in the frequency domain using diffusion models. \textit{However, these perception-centric methods often fail to improve the accuracy of high-level visual tasks and may introduce artificial colors and complex backgrounds as side effects} \cite{wangUnderwaterImageEnhancement2023}.

\subsection{UIE Datasets inspired by human visual perception}
The training of UIE algorithms requires paired data from real underwater images to enhanced ground truth. Considering the complex and dynamic underwater scenes, collection images and establishing paired groundtruth are challenging. 

Fabbri et al. \cite{fabbriEnhancingUnderwaterImagery2018} used a generative adversarial network to simulate low-quality underwater images with color casts and blur from clean images. Islam et al.  \cite{islamSimultaneousEnhancementSuperresolution2020a} constructed a simultaneous enhancing and super-resolution dataset including 1620 real-collected underwater images. Furthermore, they established the EUVP dataset \cite{islamFastUnderwaterImage2020}, which includes more scene and target differences. However, these degradation-based methods rely on training samples and can easily produce unnatural colors and textures.

Li et al. \cite{liUnderwaterImageEnhancement2020} constructed a UIE benchmark dataset namely UIEB, including 890 real underwater images and their corresponding groundtruth, which point to the best enhancement result voted by volunteers. Qi et al. \cite{qiSGUIENetSemanticAttention2022} annotated the enhanced groundtruth for SUIM \cite{islamSemanticSegmentationUnderwater2020}, which includes 1635 underwater images of various target categories. Han et al. \cite{hanUnderwaterImageRestoration2021} designed a dedicated UIE dataset for underwater coral reef observation, whose reference images were obtained by inverting water environment parameters. Peng et al. \cite{pengUShapeTransformerUnderwater2023} constructed a large scale underwater dataset to cover more scenes. 

\textit{However, these datasets are designed for UIE inspired by human visual perception, without considering the impact of enhancement on downstream tasks. }The detailed comparison of these datasets is shown in Table \ref{Table:datasetcompare}.

\begin{table*}[]\centering
\caption{Comparison of different UIE benchmark datasets. The degrade or enhance methods (Trad) refer to traditional underwater image enhancement methods, (DL) refers to deep-learning-based underwater image enhancement methods, (GB) refers to Gaussian blurring. The image source (Enc) refer to enhanced images, and (HQ) refer to high-quality underwater images}\label{Table:datasetcompare}
\scalebox{0.85}{
\begin{tabular}{l|l|l|l|l|l}
\toprule
\textbf{Dataset} & \textbf{Year} & \textbf{Img. Num.}  & \textbf{GT \& Raw Generation Methods} &\textbf{Degrade \& Enhance Method} & \textbf{Objective}\\ \midrule
Fabbri et al. \cite{fabbriEnhancingUnderwaterImagery2018} & 2018 & 7941 & Degradation from HQ images & CycleGAN  & Perception-centered \\
EUVP\cite{islamFastUnderwaterImage2020} & 2020 & 11435 & Degradation from HQ images & CycleGAN  & Perception-centered \\
UFO-120\cite{islamSimultaneousEnhancementSuperresolution2020a} & 2020 & 1620 & Degradation from HQ images & CycleGAN  + GB & Perception-centered  \\
UIEB \cite{liUnderwaterImageEnhancement2020} & 2020 & 890 & Majority-voting from Enc by human & 12 Trad methods & Perception-centered \\
SUIM-E \cite{qiSGUIENetSemanticAttention2022}  & 2022 & 1635 & Majority-voting from Enc by human & 12 Trad methods & Perception-centered \\
HICRD \cite{hanUnderwaterImageRestoration2021}  & 2022 & 2000 & Imaging parameter inversion & Trad method & Perception-centered \\
LSUI \cite{pengUShapeTransformerUnderwater2023} & 2023 & 4279 & Majority-voting from Enc by human & 12 Trad + 6 DL methods & Perception-centered \\ \midrule
TI-UIED (Ours) & 2025 & 1635 & Majority-voting from Enc by downstream tasks & 5 Trad + 4 DL methods & Downstream-task inspired \\ \bottomrule
\end{tabular}}
\end{table*}

\subsection{UIE for downstream vision tasks}
Image recognition accuracy decreases when input images are blurry, noisy, or hazy. For in-air images, there is a consensus that improving the visual quality of input images can improve the performance of downstream tasks. Consequently, research has focused on integrating enhancement tasks with high-level tasks \cite{liuImageAdaptiveYOLOObject2022} \cite{liuImprovingNighttimeDrivingScene2023} \cite{huangDSNetJointSemantic2021} \cite{liDetectionFriendlyDehazingObject2023}. \textit{However, these task-driven approaches are challenging to apply to underwater images.} First, obtaining real reference images without distortion is difficult in underwater environments. Second, the feature requirements for underwater vision tasks differ from those for in-air images. Some studies have attempted to solve this type of UIE problem. Yu et al. \cite{yuTaskFriendlyUnderwaterImage2024} proposed Task-Friendly UIE (TFUIE), which uses a disentangled representation approach to enhance content features using in-air images. Zhang et al. \cite{zhangWaterFlowHeuristicNormalizing2023} \cite{zhangHUPEHeuristicUnderwater2025} introduced HUPE, employing a detection-aware module to transfer implicit semantics into the enhancement process. Qi et al. \cite{qiSGUIENetSemanticAttention2022} presented SGUIE-Net, incorporating semantic information as high-level guidance for differential enhancement. Liu et al. \cite{liuTwinAdversarialContrastive2022a} introduced Twin Adversarial Contrastive Learning (TACL) UIE, embedding a task-aware feedback module in the enhancement process.  \textit{However, the introduction of in-air image features and human perception-friendly elements may limit their generalization across diverse underwater conditions, reducing the generalization ability of these methods to more underwater downstream tasks. }

To address this issue, we take a twofold approach:  1) constructing a UIE dataset tailored to support downstream tasks, and 2) leveraging downstream tasks as losses and priors to guide the UIE network in learning task-specific features.

\begin{figure}[t]
  \centering
  \includegraphics[width=\linewidth]{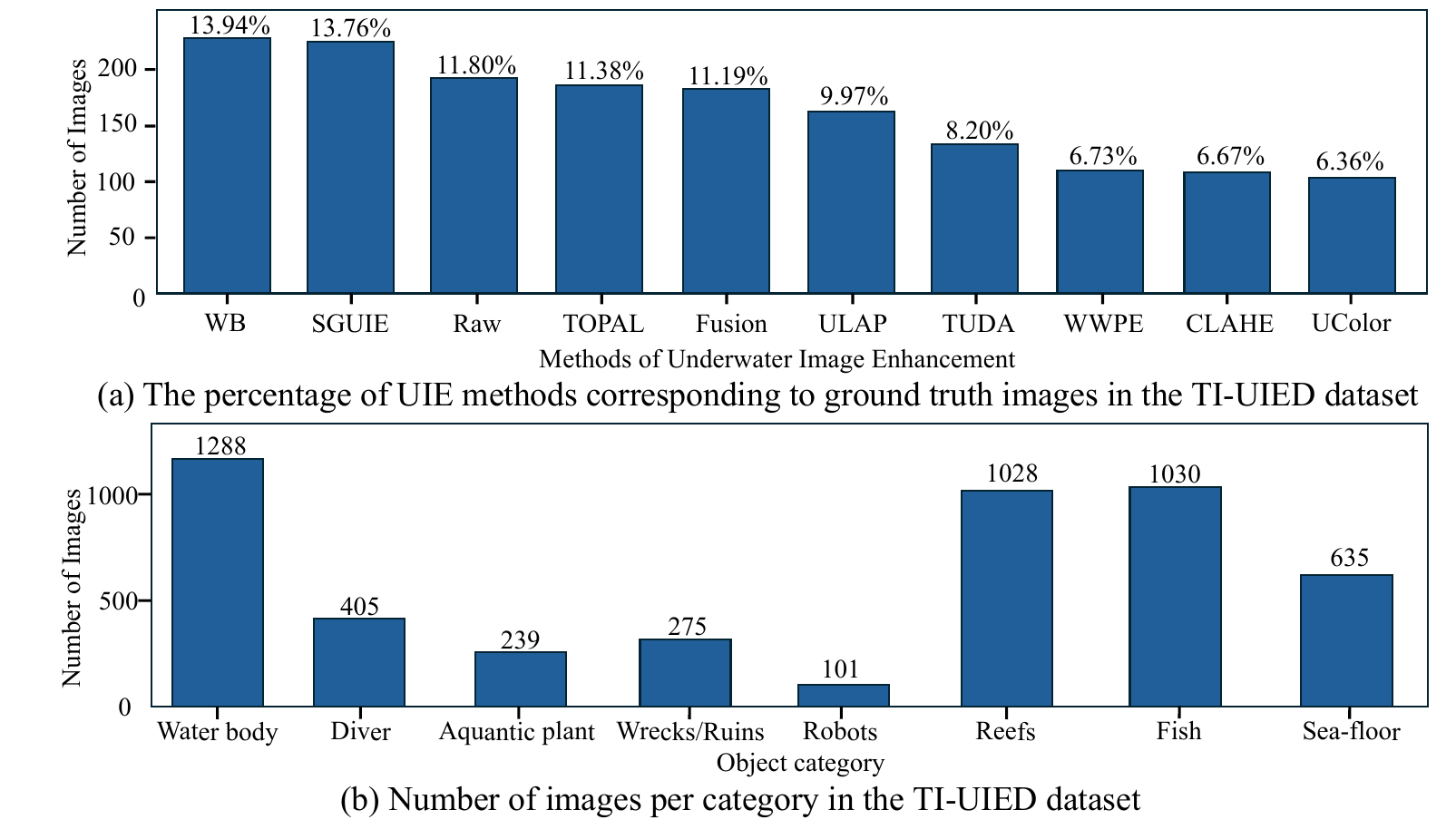}
  \caption{(a) The percentage of images from different UIE sources corresponding to ground truth in the entire TI-UIED dataset. (b) Number of images containing each object category.}
  \label{Figure_datasetcount}
\end{figure}

\begin{figure*}[t]
  \centering
  \includegraphics[width=\linewidth]{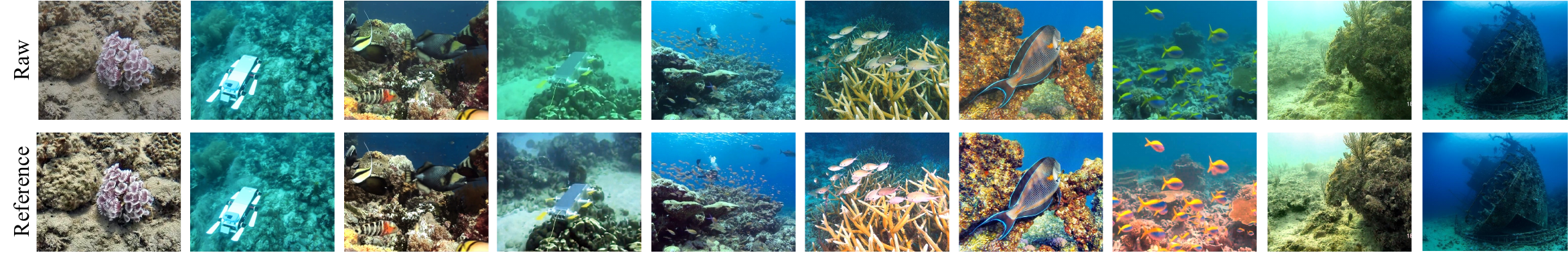}
  \caption{Example raw and references images in the TI-UIED dataset. }
  \label{Figure_datasetphoto}
\end{figure*}

\section{Proposed Task-inspired UIE Dataset}\label{Section:3}
To establish a task-inspired UIE dataset, we draw inspiration from human visual perception-centric UIE methods. In those approaches, human volunteers select the optimal enhanced image from a set of enhancement results \cite{liUnderwaterImageEnhancement2020}. The reference is determined by the majority vote of multiple volunteers. In this study, we adapt this process to simulate human visual perception using semantic segmentation methods, as illustrated in Figure \ref{Figure:overall}. 

\subsection{Raw Underwater Images Collection}
Considering the selection of reference images through semantic segmentation indicators, we build the TI-UIED dataset based on the SUIM \cite{islamSemanticSegmentationUnderwater2020} dataset. The dataset contains a total of 1635 real underwater images with pixel-level category annotations. There are eight semantic categories: water body, fish, reefs, aquatic plants, wrecks/ruins, human divers, robots, and sea-floor.  These images contain a wide variety of scenes, water types, lighting categories, and target types, making them more suitable as target images for UIE and task networks training and evaluating.

\subsection{Ground-truth Image Generation}
We attempt to use underwater task networks to simulate the “subjective” evaluations by human volunteers. Initially, two distinct semantic segmentation datasets are curated: the target set and the training set. For both datasets, a range of UIE algorithms is applied to the raw images to generate enhanced versions. To mimic human perception process, a series of semantic segmentation networks are trained on the enhanced images in the training set, enabling them to learn the transformation from enhanced images to  the corresponding semantic masks. Subsequently, the networks are tested on the target set, where each enhanced image is evaluated based on its performance.  The performance of each enhancement method is quantified using the mean Intersection over Union (mIoU) score. For each image, the enhancement method that yields the highest mIoU score  in the target set is selected as the optimal enhancement result. Finally, a majority voting mechanism is employed to determine the reference image for each raw underwater image.

Specifically, we select the SUIM dataset as the target set, and make usage of the UIIS dataset \cite{lianWaterMaskInstanceSegmentation2023} as the training set to train various semantic segmentation networks. For the UIE methods, we choose five traditional enhancement methods: CLAHE \cite{rezaRealizationContrastLimited2004}, Fusion \cite{ancutiEnhancingUnderwaterImages2012}, ULAP \cite{songRapidSceneDepth2018}, WWPE \cite{zhangUnderwaterImageEnhancement2024}, and WB \cite{ebner2021color}, as well as four deep learning-based methods: SGUIE \cite{qiSGUIENetSemanticAttention2022}, TOPAL \cite{jiangTargetOrientedPerceptual2022}, TUDA \cite{wangDomainAdaptationUnderwater2023}, and UColor \cite{liUnderwaterImageEnhancement2021}.  The source codes and pretrained models for all these methods are provided by their authors. Since existing UIE methods may not always enhance downstream task performance, we also include raw images in the comparison. If the raw image yields the best performance among all candidates, it is directly selected as the ground truth without enforcing additional enhancement. For the task networks, a number of 7 models we trained include UNet\cite{ronnebergerUnetConvolutionalNetworks2015}, Segformer\cite{xieSegFormerSimpleEfficient2021a}, PSPNet\cite{zhaoPyramidSceneParsing2017}, DeepLabV3\cite{chenRethinkingAtrousConvolution2017a}, DeepLabV3+\cite{chen2018encoder}, MaNet\cite{hettihewaMANetMultiattentionNetwork2023}, and FPN\cite{seferbekovFeaturePyramidNetwork2018}. To reduce the impact of randomness, the training process was repeated three times.  In total, we trained $(5+4+1)\times7\times3$ models. The percentage of ground truth images from the results of different UIE methods and the number of images containing each object category are shown in Figure \ref{Figure_datasetcount}. Some image examples in TI-UIED dataset are shown in \ref{Figure_datasetphoto}.

\begin{figure*}[h]
  \centering
  \includegraphics[width=0.9\linewidth]{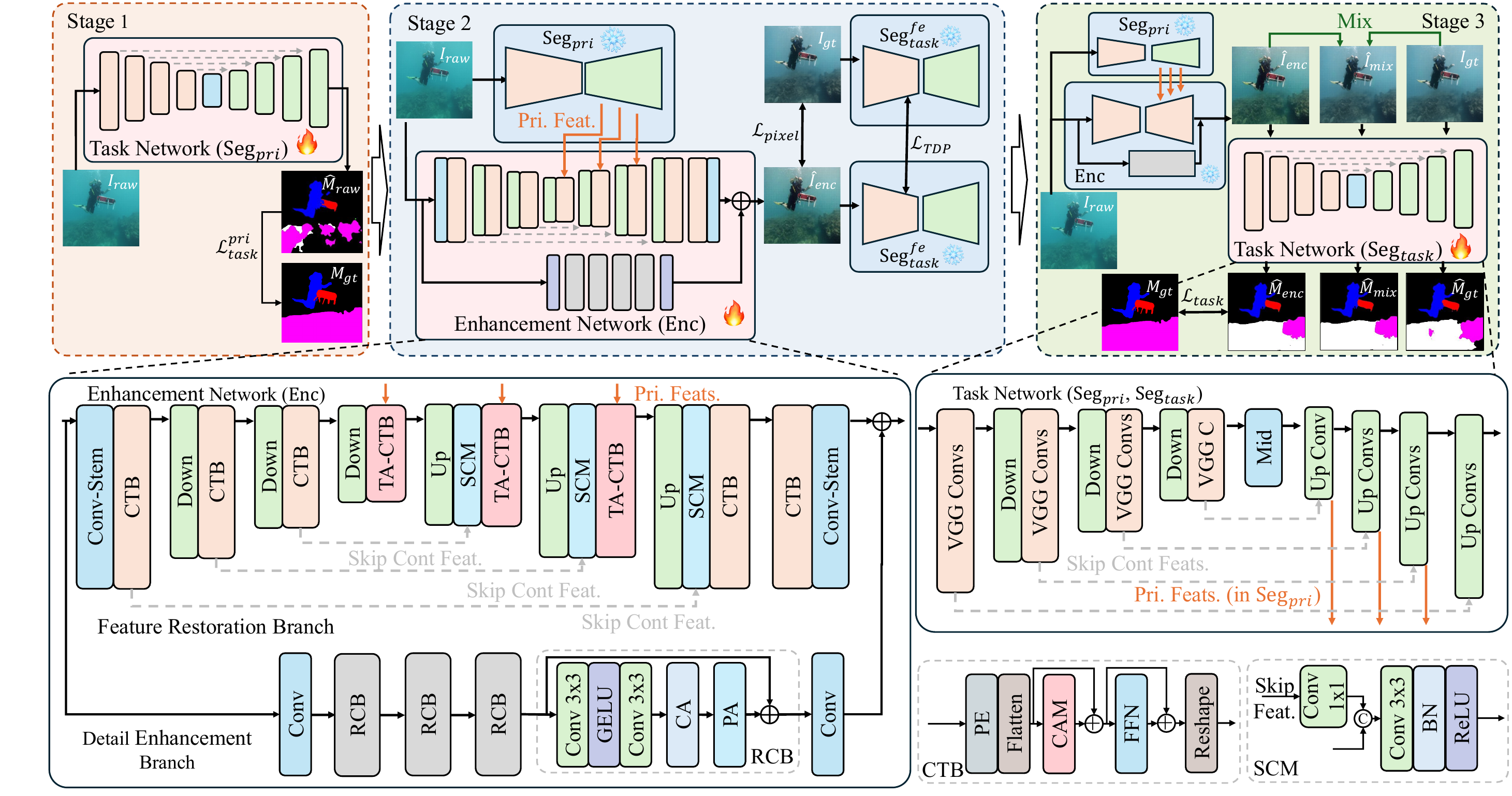}
  \caption{Overview of the proposed DTI-UIE framework, which consists of the enhancement network $\mathbf{Enc}$ and two task network $\mathbf{Seg}_{pri}$, $\mathbf{Seg}_{task}$. In the first stage, the task network $\mathbf{Seg}_{pri}$ for task-relevant prior is trained. During the second stage, DTI-UIE updates the enhancement network $\mathbf{Enc}$ using the TDP loss and pixel loss, while both two
  task networks are frozen. In the third stage, the task network $\mathbf{Seg}_{task}$ for TDP loss is updated, using different kinds of enhanced and mixed images.}\label{Fig:network}
\end{figure*}

\section{Proposed Task-inspired UIE Framework}\label{Section:4}
In this Section, we describe the structure of the proposed DTI-UIE network in detail. This specifically includes the structure of the proposed feature restoration branch, detail enhancement branch, and TA-CTB, as well as the details of the three-stage training framework and TDP loss.
\subsection {Motivation and Overview} 
\textbf{Overall Arcchitecture.}  Figure. \ref{Fig:network} illustrates the proposed DTI-UIE framework. Motivated by principles of human visual perception, particularly the complementary processing of global semantic information and local fine-grained details, DTI-UIE is designed as a two-branch enhancement network coupled with a three-stage training framework. In human perception, broad contextual understanding and detailed structural analysis are jointly integrated to support robust target recognition under complex visual conditions. Following this feature, the proposed Feature Restoration Branch (FRB) focuses on extracting and recovering global, task-relevant semantic features that are critical for downstream recognition tasks. In parallel, the Detail Enhancement Branch (DEB) operates at full spatial resolution to preserve fine textures, edges, and structural details that are easily degraded by encoder–decoder architectures but are essential for accurate detection and segmentation. The outputs of the two branches are adaptively fused to generate images that balance semantic integrity and local detail fidelity.

Furthermore, motivated by the experience-dependent modulation observed in human visual perception, TA-CTB is introduced to incorporate task-specific priors into the enhancement process. By injecting task-relevant representations derived from task networks, TA-CTB enables the enhancement network to selectively emphasize features that are most informative for recognition, thereby aligning feature restoration with downstream task requirements. To further strengthen this alignment, DTI-UIE adopts a three-stage training strategy that alternately optimizes the task and enhancement networks in a feedback-driven manner of vision perception. A learnable TDP loss is employed to guide the network toward producing representations that are not only visually enhanced but also highly compatible with task objectives. Together, the proposed architecture and training framework form a tightly coupled perception-inspired enhancement system tailored for underwater vision tasks.

\textbf{Problem definition.} Given a raw underwater image $I_{raw}\in \mathbb{R}^{H \times W \times 3}$ with width $W$ and height $H$, its corresponding enhance groundtruth $I_{gt}\in \mathbb{R}^{H \times W \times 3}$, and semantic segmentation mask $M_{gt} \in \mathbb{R}^{H \times W \times 1}$, the DTI-UIE framework can be modeled as
:
\begin{equation}
\hat{T}_{pri} =\mathbf{Seg}_{pri}(I_{raw};\theta_{pri}),
\end{equation}
where $\mathbf{Seg}_{pri}$ is a task network parameterized by $\theta_{pri}$, providing the target-relevant prior features $\hat{T}_{pri}$ These prior features are then used as input to the enhance network:
\begin{equation}
\hat{I}_{enc} =\mathbf{Enc}(I_{raw},\hat{T}_{pri};\theta_{enc}),
\end{equation}
where $\hat{I}_{enc}\in \mathbb{R}^{H \times W \times 3}$ is the enhanced result, $\mathbf{Enc}$ represents the enhancement network parameterized by $\theta_{enc}$. The goal of DTI-UIE is to generate a preprocessed image suitable for downstream tasks. These downstream tasks can be defined using a semantic segmentation network: 
\begin{equation}
\hat{M}_{enc} =\mathbf{Seg}_{task}(\hat{I}_{enc};\theta_{task}),
\end{equation}
where $\mathbf{Seg}_{task}$ is a downstream task network parameterized by $\theta_{task}$, which predicts the segmentation mask $\hat{M}_{enc}\in \mathbb{R}^{H \times W \times 1}$ from the enhance image. 

The objective of the training stage is formally defined as:
\begin{equation}
    \underset{\theta_{enc},\theta_{pri},\theta_{task}}{min} \mathcal{L}_{enc}(\hat{I}_{enc},I_{gt})+\mathcal{L}_{task}(\hat{M}_{enc},M_{gt}),
\end{equation}
where $\mathcal{L}_{enc}$ evaluates the quality of the enhanced results, and $\mathcal{L}_{task}$ assesses the performance of the downstream task.

\subsection{The Feature Restoration Branch}
UIE methods designed for human visual perception emphasize color realism, natural details, and visual optimization \cite{liTCTLNetTemplateFreeColor2024} \cite{zhouIACCCrossIlluminationAwareness2024}, which differ from the task-specific high-frequency content requirements for downstream tasks \cite{wangUnderwaterImageEnhancement2023}. To address this, we design the FRB, which focuses on recovering high-frequency features relevant to the target. The FRB uses a UNet-shaped encoder-decoder network to extract and restore image features from local to global levels. The structure of the FRB is shown in Figure \ref{Fig:network}.

Given a raw underwater image $I_{raw}$, the initial features $X_{0}$ are extracted using a convolutional stem layer with a $3 \times 3$ kernel, projecting into a 32-channel feature map. These features are then processed by Conv-attention Transformer Blocks (CTBs) \cite{xuCoScaleConvAttentionalImage2021} to restore target-related information within a UNet-shaped encoder–decoder framework. In the encoder, the FRB is composed of three hierarchical stages followed by a bottleneck stage. At each stage $i$, the input feature map $X_i \in \mathbb{R}^{H_i \times W_i \times C_i}$ is partitioned into a flattened representation $X'_i\in \mathbb{R}^{N_i \times C_i}$, where $N_i,C_i$ denote the number of tokens and embedding dimension, respectively. A convolutional attention module (CAM) is then applied, which integrates convolutional positional encoding, factorized self-attention, and convolutional relative position encoding to produce an attention-enhanced representation $A_i\in \mathbb{R}^{N_i \times C_i}$.  The output is further refined by a feed-forward module and reshaped back to their origin spatial dimensions. Then, the feature map is down-sampled via a convolution with a $1\times1$ kernel and stride of 2, producing feature maps with progressively increased channel dimensions. Specifically, the encoder generates feature representations with channel sizes $C_i \in \{32, 64, 128\}$ at resolutions $\{H \times W$, $H/2 \times W/2$, $H/4 \times W/4\}$. In the bottleneck stage, task-related priors from the task network are incorporated into the FRB by introducing semantic guidance into the CTBs, resulting in Task-Aware CTBs (TA-CTBs), resulting in feature representations with channel size of $256$ at resolutions $H/8 \times W/8$. 

In the decoder stage, three TA-CTBs followed by one CTB are connected in series, with a $2\times$ upsampling operation performed at the start of each block via transposed convolution. Skip Connection Modules (SCM) are employed to fuse the decoder features with the corresponding encoder features at the same scale. Specifically, the encoder skip features are first aligned by a $1\times1$ convolution and then concatenated with the upsampled decoder features along the channel dimension, followed by a $1\times1$ convolution with batch normalization and ReLU activation to project the fused features back to the original channel dimension. This design enables effective cross-scale feature fusion while preserving spatial details. Finally, after a CTB used as a refinement block, a $3\times3$ convolution layer reconstructs the 3-channel enhanced image, with a global residual connection added to stabilize training and preserve low-level image information. Overall, the FRB integrates multi-scale feature extraction, task-aware attention, and effective cross-scale fusion to restore task-relevant semantic structures while preserving fine spatial details, thereby providing robust and informative representations for downstream underwater vision tasks.

\begin{figure}[t]
  \centering
  \includegraphics[width=0.9\linewidth]{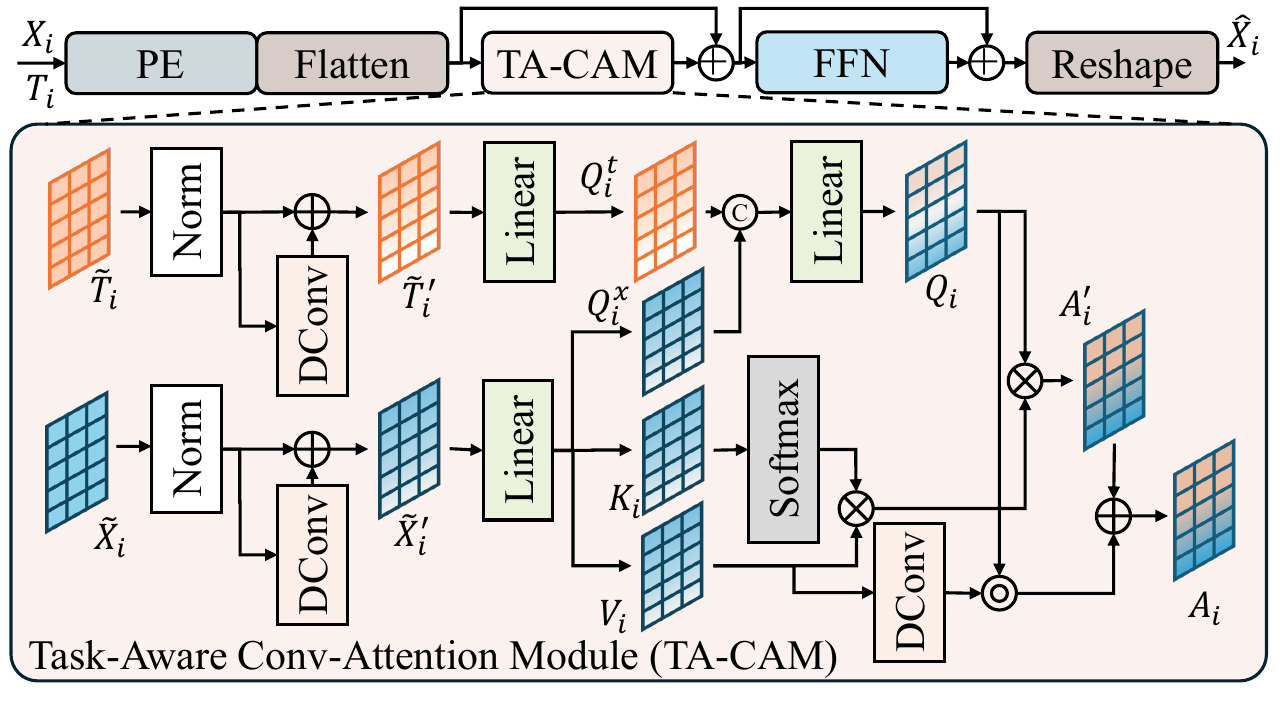}
  \caption{Structure of TA-CTB. At the $i$-th layer, the TA-CAM mixs the target-relevant prior $T_i$ and image feature map $X_i$ and produces the output feature map $\hat{X}_i$.}\label{Figure:TACTB}
\end{figure}

\subsection{Task-Aware Conv-attention Transformer Block}

In human visual perception, recognition relies on the modulation of attentional focus and selective emphasis by prior knowledge related to the task rather than being simply driven by bottom-up sensory signals. Motivated by this task-dependent modulation principle, a task-driven module that leverages priors extracted from a task network to guide the enhancement process is designed, termed the Task-Aware Conv-attention Transformer Block (TA-CTB), as illustrated in Figure \ref{Figure:TACTB}. By incorporating task-specific representations into the attention mechanism, TA-CTB enables the enhancement network to adaptively emphasize features that are more informative for downstream recognition tasks, thereby facilitating the generation of task-friendly enhanced images. We use a UNet \cite{ronnebergerUnetConvolutionalNetworks2015} with a VGG-16 \cite{simonyanVeryDeepConvolutional2014} backbone as the task network, utilizing the output features from the decoder. For image features $X_i\in \mathbb{R}^{H_i \times W_i \times C_i}$ and task-relevant priors $T_i\in \mathbb{R}^{H_i \times W_i \times C_i}$ in stage $i$, we first apply convolution layers to align their dimensions.  Then, a patch embedding operation converts them into tokens $\tilde{X_i}\in \mathbb{R}^{N_i \times C_i}$ and $\tilde{T_i}\in \mathbb{R}^{N_i \times C_i}$, where $N_i,C_i$ refer to the number of tokens and the embedding dimension, respectively. 

The TA-CTB integrates a task-aware convolutional attention module (TA-CAM) that combines target-relevant priors and enables cross-model interaction of heterogeneous task representations. In TA-CAM, given input tokens $\tilde{X_i}$ and $\tilde{T_i}$, we apply convolutional position encoding \cite{xuCoScaleConvAttentionalImage2021} to insert the position relationships into the tokens:
\begin{equation}
    \tilde{X'_i} = DConv(\tilde{X_i})+\tilde{X_i}, 
\end{equation}
\begin{equation}
    \tilde{T'_i} = DConv(\tilde{T_i})+\tilde{T_i},
\end{equation}
where $DConv$ refer to $3\times3$ depthwise convolution. This process helps capture spatial information at multiple scales. Next, we generate query ($Q_i^f=\mathrm{W}_q^f \tilde{X_i}$), key ($K_i=\mathrm{W}_k \tilde{X_i}$) and value ($V_i=\mathrm{W}_v \tilde{X_i}$) from the image tokens $\tilde{X_i}$, with $\mathrm{W}_q$, $\mathrm{W}_k$ and $\mathrm{W}_v$ as linear projection matrices. Simultaneously, we generate a query for the task-relevant tokens ($Q_i^t=\mathrm{W}_k^t \tilde{T_i}$). By applying linear projection to mix the queries, we form the mixed query $Q_i$:
\begin{equation}
    Q_i = \mathrm{W}_q^m concat(Q_i^f,Q_i^t).
\end{equation}
The factorized self-attention operation, which models long-range dependencies and spatial relationships between patches, is then applied to compute the task-aware attention map:
\begin{equation}
    A_i'=\frac{Q_i}{\sqrt{C_i}}(Softmax(K_i)^\top V_i), 
\end{equation}
where $\mathrm{W}_q^m$ is a linear projection matrix for task-aware query mixing, and $A_i \in \mathbb{R}^{N_i \times C_i}$ is the task-aware attention map, representing the interrelationship between cross-modal representations. We then apply convolutional relative position encoding to incorporate local positional information for further enhance spatial understanding between regions: 
\begin{equation}
    A_i=A_i'+Q_i\circ DConv(P_i,V_i) , 
\end{equation}
where $P_i$ is the relative position encoding and $\circ$ denotes the Hadamard product. Finally, we use the attention map $A_i$ to restore the output feature map:
\begin{equation}
    \hat{X}_i= \mathbf{FF}(A_i'+X_i)+X_i.
\end{equation}
where $\mathbf{FF}$ denotes feed-forward network for non-linearity and better feature representation. The resulting $\hat{X}_i\in \mathbb{R}^{H_i \times W_i \times C_i}$ is the final feature map of decoder $i$ which is used as input for the next decoder after a $2\times$ upsampling operation and skip connection. TA-CTB ensures that the model captures both global dependencies and fine-grained relationships with task priors, providing a robust and efficient way to learn task-relevant features.

\subsection{The Task Network}
A standard semantic segmentation model is used as the task network, implemented as an UNet-shaped \cite{ronnebergerUnetConvolutionalNetworks2015} encoder-decoder network. The encoder of the task network uses VGG-16\cite{simonyanVeryDeepConvolutional2014} feature extractor and is split into four convolutional blocks $\{T^E_1,\dots,T^E_4\}$, which progressively extract hierarchical representations from the input image. Each block consists of stacked $3\times3$ convolutions with batch normalization and ReLU. The decoder follows the UNet style design with four up-sampling blocks $\{T^D_1,\dots,T^D_4\}$. Each stage performs $2\times$ up-sampling via transposed convolution and concatenates the feature with the corresponding encoder feature through skip connections. The features are then refined by two $3\times3$ convolutions, each followed by batch normalization and ReLU activation. A final $1\times1$ convolution produces the prediction map.

Serving as task-relevant priors, the task network provides multi-scale decoder features $\{F^T_4,\dots,F^T_1\}$. The features $\{F^T_4,F^T_3,F^T_2\}$ are injected into the enhancement network via TA-CTB to guide task-friendly enhancement, and the feature map $F^T_1$ is further used to construct task-driven perceptual supervision via the TDP loss.

\subsection{The Detail Enhancement Branch}
Human perception-inspired UIE may introduce noise, edge blur, and texture degradation \cite{wangUnderwaterImageEnhancement2023}. While these artifacts have minimal impact on human vision, they can significantly reduce the accuracy of downstream tasks. Since the encoder-decoder structure tends to lose structural and spatial details \cite{qiSGUIENetSemanticAttention2022} \cite{zhouHCLRNetHybridContrastive2024}, we design an independent DEB that operates at the raw image resolution to preserve fine textures.

As shown in Figure \ref{Fig:network}, the proposed DEB consists of four residual convolution blocks (RCB) \cite{heDeepResidualLearning2016}. Each RCB integrates convolutions, a CAM, and a pixel attention module (PAM) \cite{wooCBAMConvolutionalBlock2018}.  The CAM captures global channel dependencies through global max and average pooling, aiding in the selective recovery and fusion of key features from high-resolution images. The PAM assigns attentional weights to each pixel, enhancing spatial features and details, and generating images that are more effective for downstream visual tasks.

\subsection{Task-Driven Perceptual Loss and Multi-stage Training Framework}
Finally, inspired by the iterative and feedback-driven nature of human visual perception, the proposed DTI-UIE adopts a three-stage training framework. Specifically, in human vision, perceptual interpretation and task understanding are progressively refined through interaction. Inspired by this, the designed three-stage training framework alternately optimizes the task network and the enhancement network, allowing task-relevant priors and enhancement representations to be mutually refined. Meanwhile, a learnable task-driven perceptual (TDP) loss is introduced to explicitly align the enhancement outputs with downstream task objectives, guiding the enhancement network toward producing task-friendly representations.

In the first stage of network training, we extract task-relevant prior knowledge $\hat{T}_{pri}$ from raw underwater images $I_{raw}$ using the prior task network $\mathbf{Seg}_{pri}$. To prevent the $\mathbf{Seg}_{pri}$ from providing shortcut features and improve its robustness, we dynamically train this task network. Since we use a semantic segmentation network as the prior task network, the task-specific loss $\mathcal{L}_{task}^{pri}$ is defined as:
\begin{equation}
    \mathcal{L}_{task}^{pri}=CE(\mathbf{Seg}_{pri}(I_{raw}),M_{gt}),
\end{equation}
where $CE(\cdot)$ is the cross-entropy loss for segmentation task.

In the second stage, we train the enhancement network $\mathbf{Enc}$ using the raw images $I_{raw}$, the corresponding enhanced ground trurh $I_{gt}$, and target-relevant prior $\hat{T}_{pri}$ provided by $\mathbf{Seg}_{pri}$. For the enhanced images, we first apply the pixel loss:
\begin{equation}
    \mathcal{L}_{pixel}=MSE(\mathbf{Enc}(I_{raw}, \hat{T}_{pri}),I_{gt}). 
\end{equation}
However, since the goal of DTI-UIE is to provide preprocessed images beneficial for downstream tasks, the traditional pixel-based loss cannot establish feature-level associations with these tasks, making the enhanced results less useful. Inspired by \cite{kimImageSuperResolutionImage2024}, we introduce a trainable task network to generate a TDP loss for the enhancement results (specifically, a UNet segmentation network with a VGG-16 backbone). The TDP loss aims to maximize the similarity between the enhanced images $\hat{I}_{enc}$ and the ground-truth $I_{gt}$ within the feature extraction latent space of the task network, ensuring the enhancement results are closely related to the task. We define the TDP loss as:
\begin{equation}
    \mathcal{L}_{TDP}=\left \| \mathbf{Seg}_{task}^{fe}(\hat{I}_{enc})-\mathbf{Seg}_{task}^{fe}(I_{gt})\right \| _1,
\end{equation}
where $\mathbf{Seg}_{task}^{fe}$ refers to the feature extractor of the downstream task network $\mathbf{Seg}_{task}$. Specifically, we calculate the feature differences of the last encoder. At the second stage, both the $\mathbf{Seg}_{pri}$ and the $\mathbf{Seg}_{task}$ remain frozen.

The purpose of the third stage is to update the task network $\mathbf{Seg}_{task}$, enabling it to learn more comprehensive task-relevant features and avoid biased representations, thereby improving the effectiveness of the TDP loss. To achieve this, we utilize three types of images, $\hat{I}_{enc}$, $I_{gt}$, and $\hat{I}_{mix}$, as training inputs. These images are used to guide the perceptual task network toward better feature representation. Among them, the mix image $\hat{I}_{mix}$ is a new image created by randomly mixing $I_{gt}$ and $\hat{I}_{enc}$ at the patch level. Specifically, a mixing mask from a randomly initialized tensor is created, which is then passed through a sigmoid activation and scaled via bilinear interpolation to match the spatial dimensions of the enhanced images. The resulting mask is used to blend $\hat{I}_{enc}$ and $I_{gt}$ at a pixel level to create $\hat{I}_{mix}$, where the mixing proportion is determined by the mask values. This approach aims to encourage the network to learn more robust task-relevant features by forcing it to handle both clean and enhanced regions within a single image, thus improving generalization. The training loss is defined as:
\begin{equation}
    \mathcal{L}_{task}=\textstyle \sum CE(\mathbf{Seg}_{task}(\tilde{I}),M_{gt}), \tilde{I}\in\{\hat{I}_{enc},I_{gt},\hat{I}_{mix}\}.
\end{equation}

In summary, the objectives of the DTI-UIE in each stage $(\mathrm{S})$ are defined as follows:
\begin{equation}
\begin{matrix}
\underset{\theta_{pri}}{min} \ \mathcal{L}_{task}^{pri}(\mathbf{Seg}_{pri}(I_{raw}),M_{gt}), \mathrm{in \ S1,}
 \\
\underset{\theta_{enc}}{min} \ \mathcal{L}_{pixel}(\mathbf{Enc}(I_{raw},\hat{T}_{pri}),I_{gt}) + \mathcal{L}_{TDP}, \mathrm{in \ S2,}
 \\
\underset{\theta_{task}}{min} \ \textstyle \sum\mathcal{L}_{task}(\mathbf{Seg}_{task}(\tilde{I}),M_{gt}), \tilde{I}\in\{\hat{I}_{enc},I_{gt},\hat{I}_{mix}\}, \mathrm{in \ S3.}
\end{matrix}
\end{equation}

\begin{figure*}[t]
  \centering
  \includegraphics[width=\linewidth]{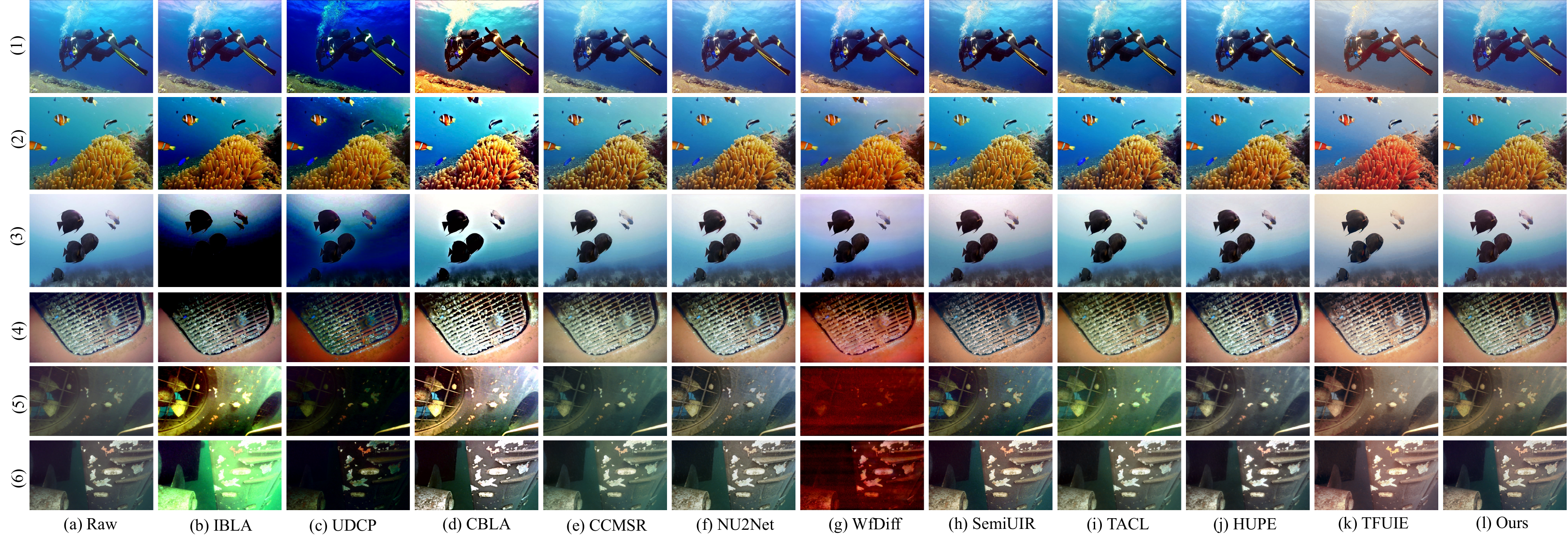}
  \caption{Visual comparison with different UIE methods on SUIM \cite{islamSemanticSegmentationUnderwater2020}  (1-3) and LIACi \cite{waszakSemanticSegmentationUnderwater2022} (4-6) dataset.}
  \label{Figure:encimages}
\end{figure*}

\section{Experiments}\label{Section:5}
In this section, we assess the effectiveness and generalization of the proposed DTI-UIE method as image preprocessing methods in improve the performance of three kinds of downstream image recognition tasks: semantic segmentation, instance segmentation, and object detection.
 
\subsection{Implementation Details}
We trained the proposed DTI-UIE network using the TI-UIED dataset and its corresponding semantic segmentation masks. All images are resized to 256×256 pixels as input. We adopt the AdamW optimizer and apply a cosine learning rate schedule starting from an initial value of 1e-4. Once trained, the DTI-UIE network is used as a preprocessing module to enhance images from task-specific datasets, which are then used to retrain various downstream networks. The performance is evaluated through statistical analysis of relevant metrics of the downstream networks. For all experiments, we strictly follow the original train and test splits of the datasets. When SUIM is used to construct TI-UIED, only the training split of the original SUIM dataset is involved in dataset construction and enhancement model training, while evaluation is performed on the original test split of SUIM dataset. Cross-task and cross-dataset evaluations are further conducted to assess generalization.

We compare DTI-UIE with several UIE methods, including traditional approaches (UDCP \cite{drewsUnderwaterDepthEstimation2016}, IBLA \cite{pengUnderwaterImageRestoration2017}, and CBLA \cite{jhaCBLAColorBalancedLocally2024}), deep learning-based methods (NU2Net \cite{guoUnderwaterRankerLearn2022}, SemiUIR \cite{huangContrastiveSemisupervisedLearning2023}, WfDiff \cite{zhaoWaveletbasedFourierInformation2024}, and CCMSR \cite{qiDeepColorCorrectedMultiscale2024}), and task-oriented UIE methods (TACL \cite{liuTwinAdversarialContrastive2022a}, HUPE \cite{zhangHUPEHeuristicUnderwater2025}, and TFUIE \cite{yuTaskFriendlyUnderwaterImage2024}). For all methods, we use the official implementations and pretrained weights provided by authors. All methods are implemented in PyTorch and trained on an NVIDIA RTX 3090Ti GPU.

\begin{figure*}[t]
  \centering
  \includegraphics[width=0.95\linewidth]{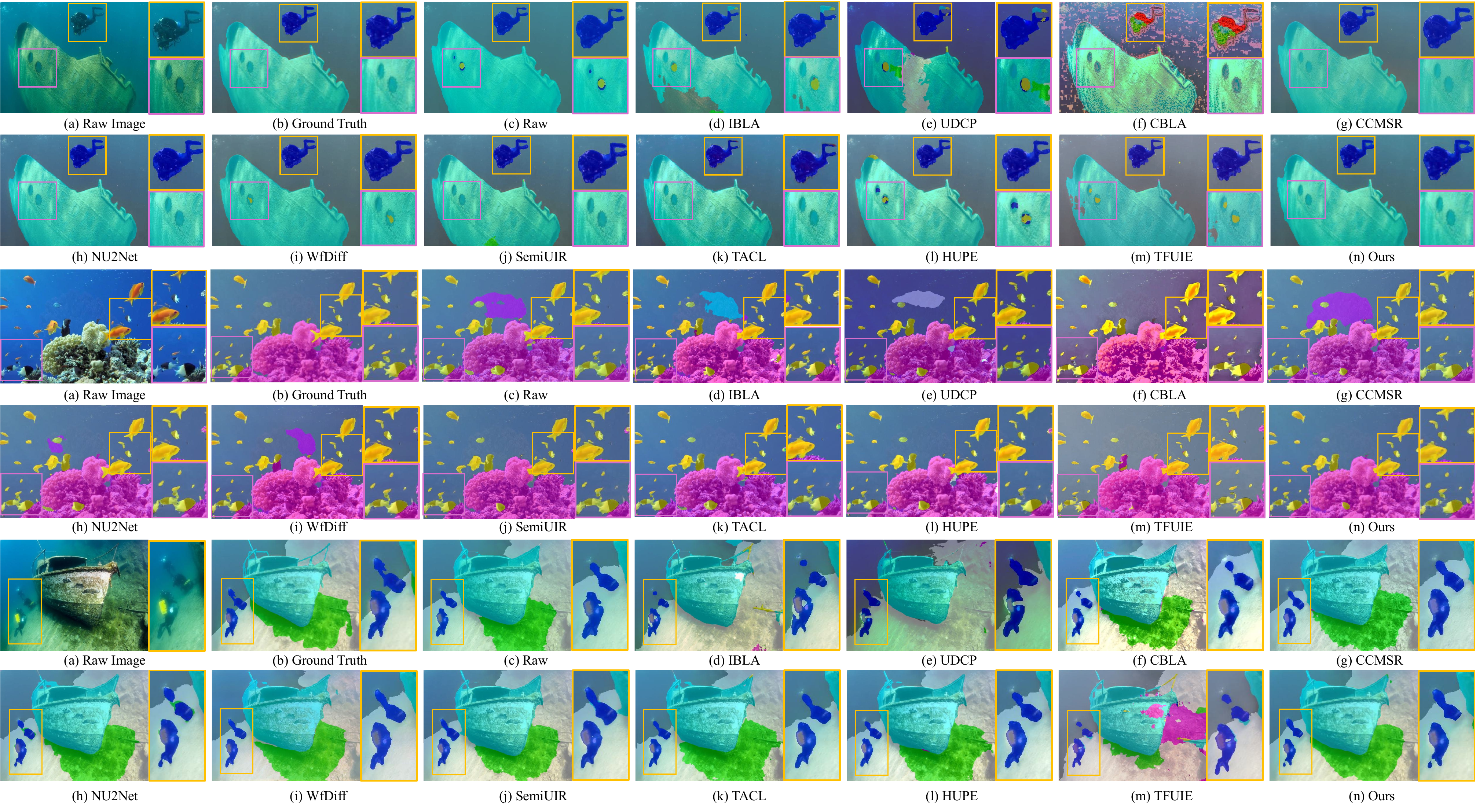}
  \caption{Segmentation results on SUIM \cite{islamSemanticSegmentationUnderwater2020} dataset. The results are obtained by training the UNet\cite{ronnebergerUnetConvolutionalNetworks2015} on raw and different enhanced images.}
  \label{Figure:suimsegresult}
\end{figure*}

\begin{figure*}[t]
  \centering
  \includegraphics[width=0.95\linewidth]{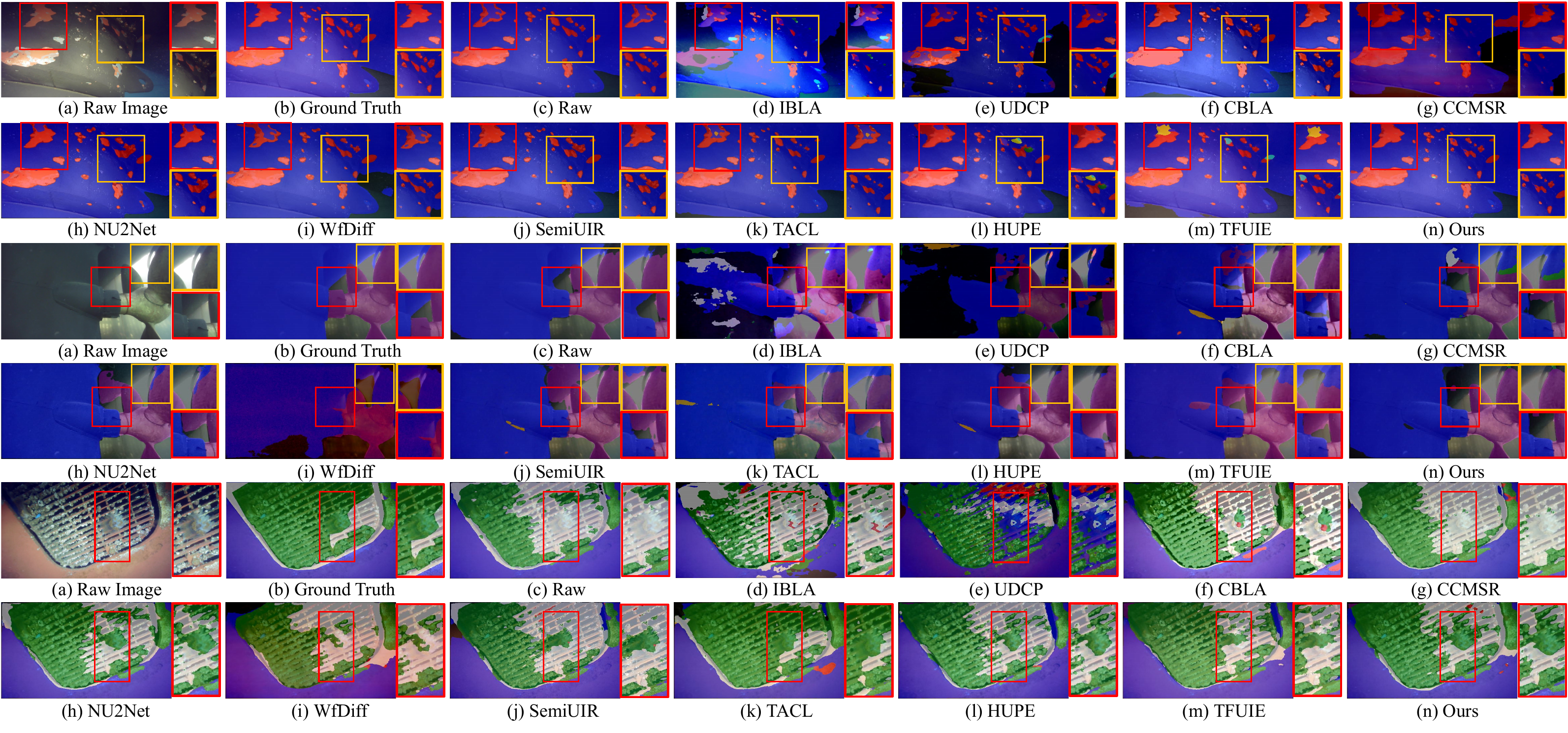}
  \caption{Segmentation results on LIACi \cite{waszakSemanticSegmentationUnderwater2022} dataset. The results are obtained by training the UNet\cite{ronnebergerUnetConvolutionalNetworks2015} on raw and different enhanced images.}
  \label{Figure:liacisegresult}
\end{figure*}

\subsection{Comparison on Semantic Segmentation}
To evaluate the performance improvement of the proposed DTI-UIE as preprocessing modules for semantic segmentation, we trained UNet \cite{ronnebergerUnetConvolutionalNetworks2015}, DeepLabV3 \cite{chenRethinkingAtrousConvolution2017a}, FPN \cite{seferbekovFeaturePyramidNetwork2018} and DPT \cite{ranftlVisionTransformersDense2021} using both enhanced and raw underwater images. This comparison was conducted on two datasets: SUIM \cite{islamSemanticSegmentationUnderwater2020}, which focuses on underwater natural image segmentation, and LIACi \cite{waszakSemanticSegmentationUnderwater2022}, which targets underwater inspection tasks for robots, specifically hull surface defect segmentation. Performance is analyzed using the Dice score and mIoU score \cite{islamSemanticSegmentationUnderwater2020}.

\begin{table*}[ht!]
\caption{Comparison for semantic segmentation in terms of mIoU($\uparrow$) and Dice($\uparrow$) on SUIM \cite{islamSemanticSegmentationUnderwater2020} and LIACi \cite{waszakSemanticSegmentationUnderwater2022} dataset. The best and second result are marked in \textbf{bold} and \underline{underline}.} \label{Table:segresult}
\centering
\setlength{\tabcolsep}{4.5pt}
\scalebox{0.8}{
\begin{tabular}{c|cc|cc|cc|cc|cc|cc|cc|cc}
\hline\toprule
\multirow{4}{*}{Method} & \multicolumn{8}{c|}{SUIM Dataset \cite{islamSemanticSegmentationUnderwater2020}} & \multicolumn{8}{c}{LIACi Dataset \cite{waszakSemanticSegmentationUnderwater2022}}\\
\cmidrule{2-17} 
& \multicolumn{2}{c|}{UNet\cite{ronnebergerUnetConvolutionalNetworks2015}} & \multicolumn{2}{c|}{DeepLabv3\cite{chenRethinkingAtrousConvolution2017a}} & \multicolumn{2}{c|}{FPN\cite{seferbekovFeaturePyramidNetwork2018}} & \multicolumn{2}{c|}{DPT\cite{ranftlVisionTransformersDense2021}} & \multicolumn{2}{c|}{UNet\cite{ronnebergerUnetConvolutionalNetworks2015}} & \multicolumn{2}{c|}{DeepLabv3\cite{chenRethinkingAtrousConvolution2017a}} & \multicolumn{2}{c|}{FPN\cite{seferbekovFeaturePyramidNetwork2018}} & \multicolumn{2}{c|}{DPT\cite{ranftlVisionTransformersDense2021}}\\
\cmidrule{2-17}
& Dice & mIoU & Dice  & mIoU  & Dice  & mIoU & Dice  & mIoU & Dice & mIoU & Dice  & mIoU  & Dice  & mIoU  & Dice  & mIoU \\
\midrule  
Raw & 72.29 & 66.52 & 78.02 & 66.51 & 74.96 & \underline{64.74} & 82.48 & \underline{73.18} & 75.55 & 54.43 & 82.51 & 55.35 & \underline{81.63} & \underline{53.66} & 81.11 & \underline{56.88} \\
UDCP \cite{drewsUnderwaterDepthEstimation2016} & 69.58 & 62.12 & 77.45 & 62.28 & 73.84 & 60.70 & 82.08 & 68.69 & 76.80 & 52.21 & 80.64 & 53.87 & 80.44 & 50.35 & 81.89 & 53.19 \\
IBLA \cite{pengUnderwaterImageRestoration2017}& 72.09 & 61.67 & 78.10 & 61.81 & 76.27 & 62.71 & 81.22 & 68.91 & 76.43 & 49.18 & 79.75 & 50.25 & 78.74 & 45.89 & 80.12 & 50.94 \\
CBLA\cite{jhaCBLAColorBalancedLocally2024}     & 72.16 & 63.63 & 75.04 & \underline{67.02}  & 75.07 & 62.79  & 81.82 & 70.73 & 77.85 & 54.15 & 82.67 & 55.22 & 79.81 & 50.95 & 79.00 & 55.30 \\ 
NU2Net\cite{guoUnderwaterRankerLearn2022}   & 73.31 & 66.62 & 78.44 & 66.63 & 76.34 & 63.76 & \underline{83.55} & 70.25 & 76.65 & \underline{54.90} & 82.47 & 55.12 & 80.23 & 52.71 & 81.52 & 56.84\\
WfDiff\cite{zhaoWaveletbasedFourierInformation2024}   & 72.37 & 66.09 & 77.62 & 64.84 & 76.12 & 64.58 & 80.56 & 70.62 & 75.75 & 49.55 & 80.38 & 51.14 & 77.27 & 46.94 & 80.36 & 51.45\\ 
SemiUIR\cite{huangContrastiveSemisupervisedLearning2023}  & 73.22 & 66.25 & 76.77 & 64.48 & 76.31 & 63.74 & 78.79 & 70.60 & 77.45 & 54.36 & 82.18 & 55.23 & 80.80 & 52.76 & 81.23 & 56.58 \\ 
CCMSR\cite{qiDeepColorCorrectedMultiscale2024} & 72.58 & 65.47 & 78.56 & 65.12 & 76.34 & 63.59 & 81.23 & 72.22 & 77.79 & 53.57 & 82.47 & 54.80 & 79.31 & 51.66 & 80.72 & 56.02 \\ \midrule
HUPE\cite{zhangHUPEHeuristicUnderwater2025}     & 69.93 & 65.89 & \underline{78.71} & 65.20 & \underline{76.64} & 64.92 & \textbf{84.09} & 72.91 & 77.65 & 54.87 & \underline{83.12} & 55.29 & 81.09 & 52.46 & 81.72 & 55.96 \\
TFUIE\cite{yuTaskFriendlyUnderwaterImage2024}    & \underline{73.51} & 64.22 & 75.26 & 64.33 & 76.34 & 61.88 & 82.30 & 68.72 & 77.50 & 52.90 & 81.52 & \textbf{55.53} & 80.24 & 52.14 & \underline{82.72} & 55.69 \\
TACL\cite{liuTwinAdversarialContrastive2022a}     & 73.13 & \underline{66.84} & 76.37 & 65.63 & 75.74 & 62.87 & 81.97 & 71.57 & \underline{77.92} & 54.21 & 82.26 & 55.42 & 79.76 & 52.84 & 80.58 & 56.84  \\ \midrule
DTI-UIE (Ours)     & \textbf{74.48} & \textbf{67.61} & \textbf{78.93} & \textbf{67.61} & \textbf{76.85} & \textbf{66.22} & 80.64 & \textbf{73.55} & \textbf{78.02} & \textbf{55.27} & \textbf{83.13} & \underline{55.47} & \textbf{81.68} & \textbf{53.72} & \textbf{82.82} & \textbf{57.17} \\ 
\bottomrule\hline
\end{tabular}}
\end{table*}

Table \ref{Table:segresult} presents the quantitative results for the semantic segmentation task. On the SUIM dataset, the proposed DTI-UIE significantly improves the Dice scores to 74.48, 78.93, 76.85, and 80.64 for the UNet, DeepLabV3, FPN, and DPT networks, respectively. A similar trend is observed in the mIoU scores, reaching 67.61, 67.61, 66.22, and 73.55 in the four networks, respectively. These results clearly demonstrate the effectiveness of incorporating target-specific knowledge during enhancement, which contributes to the performance boost of the proposed DTI-UIE as preprocessing method for downstream segmentation tasks. Notably, UIE methods designed specifically for human visual perception, such as SemiUIR and CCMSR, fail to improve downstream task accuracy and may even reduce performance compared to raw images. Traditional underwater image enhancement methods also fail to achieve better downstream task metrics. Methods that account for downstream tasks, such as HUPE and TACL, mitigate this issue to some extent. However, the proposed DTI-UIE consistently outperforms these approaches.

Experiments on the LIACi dataset further validate the effectiveness of the DTI-UIE. Compared to natural scenes, the water environment in this task is more complex, and high-frequency details are more likely to be lost. In this context, DTI-UIE achieves significant improvements, with mIoU scores of 78.02, 83.13, 81.68, and 82.82 for these four networks. These results confirm that the proposed method successfully restores task-relevant details, which are beneficial for downstream tasks performed by underwater robots.
 
Visual comparisons of some enhanced images are shown in the Figure \ref{Figure:encimages}, which shows that the proposed DTI-UIE method is able to better restore the details and textures of the target area of the image. Further, we provide visual results of segmentation mask corresponding to different UIE methods, shown in Figure \ref{Figure:suimsegresult} and Figure \ref{Figure:liacisegresult}.  The results show that the UNet trained with DTI-UIE achieves masks similar to the ground-truth, indicating that DTI-UIE is able to preserve more underlying semantic information for the segmentation network.

\subsection{Comparison on Object Detection}
For the object detection task, we employed three widely used detection models to examine the effectiveness of different UIE networks when applied as preprocessing methods, including Faster R-CNN \cite{renFasterRCNNRealTime2016}, SSD \cite{liuSSDSingleShot2016}, and DINO \cite{zhangDINODETRImproved2022}. Detection performance was evaluated using the standard mean Average Precision (mAP) metric. Specifically, we report $AP$, $AP_{50}$, $AP_{75}$, $AP_{small}$, $AP_{medium}$, and $AP_{large}$, which correspond to mAP at IoU thresholds of [0.50:0.95], 0.50, and 0.75, as well as for objects of small, medium, and large sizes, respectively.

\begin{figure*}[t]
  \centering
  \includegraphics[width=0.95\linewidth]{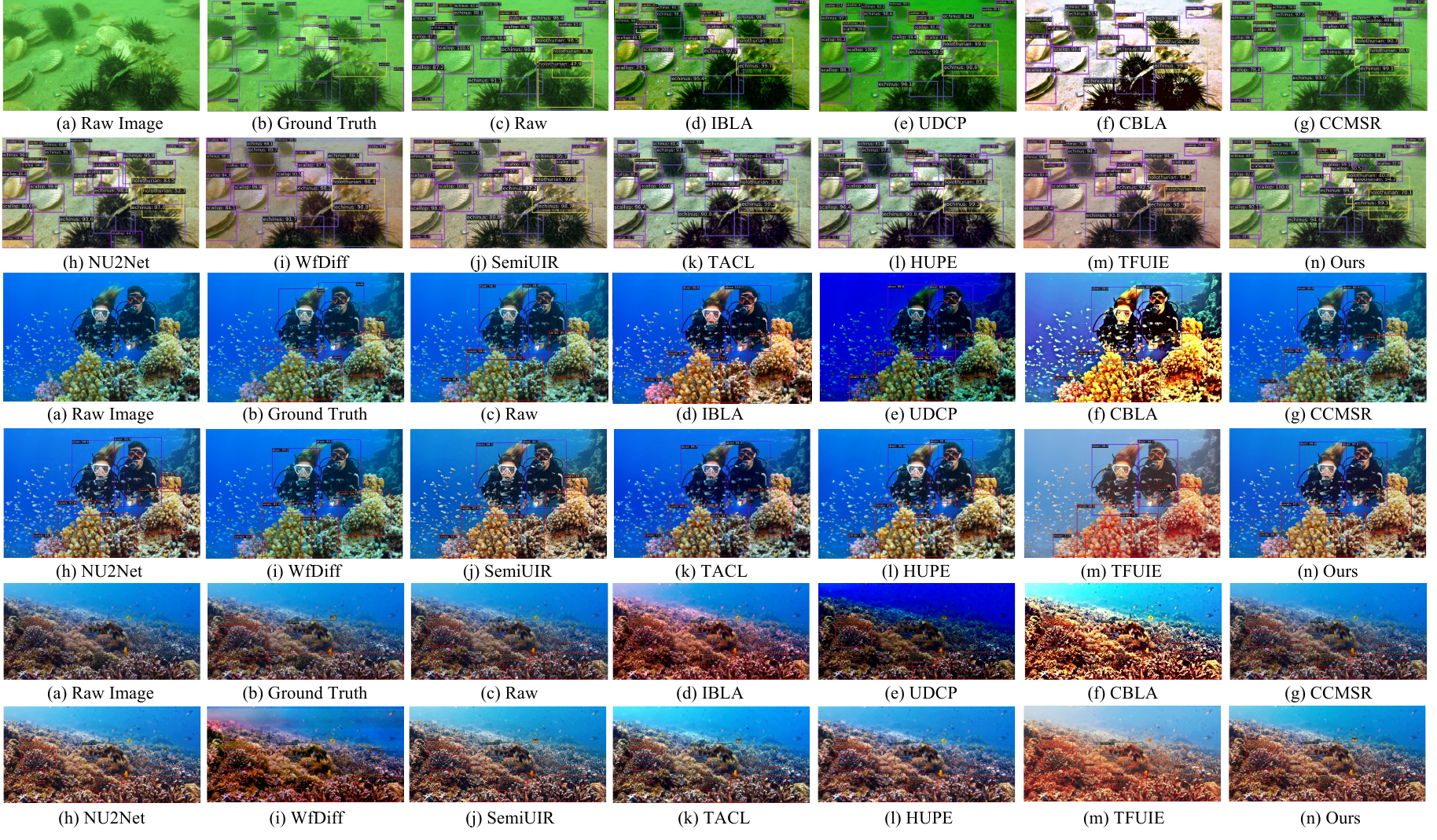}
  \caption{Visualization results of object detection on RUOD dataset\cite{fuRethinkingGeneralUnderwater2023}. The results are obtained by training the SSD \cite{liuSSDSingleShot2016} network on the raw image and different enhanced images.}
  \label{Figure:detresult}
\end{figure*}

Table \ref{Table:detresult} presents the quantitative results on the RUOD \cite{fuRethinkingGeneralUnderwater2023} dataset. The proposed DTI-UIE method consistently achieves the highest AP score compared to all other UIE methods. Specifically, DTI-UIE outperforms the Faster R-CNN model trained on SemiUIR, as well as the second-best model NU2Net, with mAP score increases of +0.50 and +0.60, respectively. Traditional underwater image enhancement methods and downstream-oriented enhancement methods have failed to achieve better results. The results on the SSD and DINO model also reflect the improvement of indicators. We further visualize the object detection results as shown in Figure \ref{Figure:detresult}. From the visualization results, it can be seen that the SSD network trained with DTI-UIE can better avoid the problem of missing detection of small targets and occluded targets, and shows a higher detection confidence in the detected targets.

\begin{table*}[]
\caption{Comparison for object detection in terms of mAPs($\uparrow$), mAPm($\uparrow$) mAPl($\uparrow$), mAP50($\uparrow$), mAP75($\uparrow$) and mAP($\uparrow$) on RUOD\cite{fuRethinkingGeneralUnderwater2023} dataset. The best and second result are marked in bold and \underline{underline}.}\label{Table:detresult}
\centering
\setlength{\tabcolsep}{2pt} 
\scalebox{0.85}{
\begin{tabular}{c|cccccc|cccccc|cccccc}
\hline\toprule
\multirow{3}{*}{Method} & \multicolumn{6}{c|}{Faster-RCNN \cite{renFasterRCNNRealTime2016}} & \multicolumn{6}{c|}{SSD \cite{liuSSDSingleShot2016}} & \multicolumn{6}{c}{DINO \cite{zhangDINODETRImproved2022}}   \\
\cmidrule{2-19} 
& mAPs & mAPm  & mAPl & mAP50 & mAP75  & mAP   & mAPs & mAPm  & mAPl & mAP50  & mAP75 & mAP & mAPs & mAPm  & mAPl & mAP50  & mAP75 & mAP   \\ 
\midrule
Raw      & 14.10 & \textbf{40.70} & 57.30 & 81.60 & 57.40  & 52.10   & \textbf{15.10} & \underline{35.80} & \textbf{55.90} & \underline{80.80} & \underline{54.30} & \underline{50.50} & 19.30 & \textbf{43.70} & \underline{64.80} & \textbf{83.60} & \underline{63.50} & \underline{59.00} \\
UDCP \cite{drewsUnderwaterDepthEstimation2016} & \underline{16.80} & 36.90 & 53.80 & 79.30 & 52.70 & 48.80 & 11.90 & 34.10 & 54.50 & 79.60 & 52.60 & 49.30 & 19.00 & 42.70 & 63.70 & 83.10 & 62.10 & 58.00 \\
IBLA \cite{pengUnderwaterImageRestoration2017} & 14.40 & 35.60 & 52.40 & 77.90 & 50.50 & 47.40 & 14.00 & 33.10 & 53.30 & 78.60 & 51.00 & 48.20 & 18.00 & 39.20 & 60.30 & 79.70 & 58.20 & 54.60 \\
CBLA\cite{jhaCBLAColorBalancedLocally2024} & 13.80 & 37.90 & 56.30 & 80.40 & 55.40  & 50.80  & 12.60 & 30.90 & 51.10 & 77.10   & 48.20    & 46.00 & \underline{19.40} & 40.50 & 62.60 & 60.60 & 81.50 & 56.60\\ 
NU2Net\cite{guoUnderwaterRankerLearn2022}   & 16.00 & 39.70 & \underline{58.00} & 82.00 & \underline{57.60}  & 52.40  & 8.20 & 34.00 & 53.80 & 79.80 & 50.90 & 48.50 & 18.40 & 43.00 & 64.40 & \underline{83.10} & 62.90 & 58.30\\ 
WfDiff\cite{zhaoWaveletbasedFourierInformation2024} & 12.00 & 37.50 & 54.60 & 79.50 & 54.20  & 49.90  & 8.50 & 32.20 & 49.00 & 75.70 & 46.90 & 45.10 & 18.30 & 39.80 & 58.80 & 78.80 & 57.80 & 54.00 \\ 
SemiUIR\cite{huangContrastiveSemisupervisedLearning2023}  & 16.00 & 40.20 & 57.90 & \underline{82.20} & 57.40  & \underline{52.50} & 9.40 &34.00 & 53.20 & 79.70 & 51.00 & 48.20 & 18.10 & 41.10 & 62.10 & 81.50 & 60.60 & 56.30\\ 
CCMSR\cite{qiDeepColorCorrectedMultiscale2024} & 16.70 & 38.30 & 55.20 & 80.50 & 54.20  & 50.10 & 10.30 & 34.70 & \underline{53.90} & 80.30 & 52.20 & 48.80 & 15.30 & 40.80 & 62.80 & 81.90 & 61.40 & 56.90 \\ \midrule
HUPE\cite{zhangHUPEHeuristicUnderwater2025} & 15.90 & 40.10 & 53.70 & 78.30 & 54.30  & 49.70  & 9.30 & 34.60 & 50.20 & 76.20 & 49.00 & 46.20 & 17.90 & 42.20 & 58.80 & 78.20 & 58.20 & 54.40 \\
TFUIE\cite{yuTaskFriendlyUnderwaterImage2024} & 10.10 & 37.50 & 39.80 & 65.80 & 45.00  & 41.50 & 8.50 & 33.10 & 53.00 & 79.10 & 50.20 & 47.60 & 14.60 & 33.80 & 54.00 & 73.80 & 51.60 & 48.40 \\
TACL\cite{liuTwinAdversarialContrastive2022a} & 11.00 & 38.00 & 56.00 & 79.90 & 55.60  & 50.90  & 12.40 & 33.00 & 52.90 & 78.90 & 50.60 & 47.80 & \textbf{20.80} & 39.70 & 62.20 & 81.40 & 60.70 & 56.30 \\ \midrule
DTI-UIE (Ours)     & \textbf{16.90} & \underline{40.50} & \textbf{58.20} & \textbf{82.50} & \textbf{58.30}  & \textbf{53.00} & \underline{14.80} & \textbf{36.20} & \textbf{55.90}  & \textbf{81.00}   & \textbf{54.80}    & \textbf{50.70} & 18.00 & \underline{43.10} & \textbf{65.30} & \textbf{83.60} & \textbf{64.00} & \textbf{59.20} \\ 
\bottomrule\hline
\end{tabular}}
\end{table*}

\subsection{Comparison on Instance Segmentation}
Additionally, we compared the UIE methods on downstream instance segmentation tasks using the UIIS dataset \cite{lianWaterMaskInstanceSegmentation2023}. Performance is analyzed using the standard mask AP metric \cite{linMicrosoftCOCOCommon2014}, including $mAP$, $AP_{50}$, $AP_{75}$, $AP_{small}$, $AP_{medium}$, and $AP_{large}$. The algorithms used for comparison include Mask R-CNN \cite{renFasterRCNNRealTime2016} and Water Mask R-CNN \cite{lianWaterMaskInstanceSegmentation2023}.

\begin{figure*}[t]
  \centering
  \includegraphics[width=0.75\linewidth]{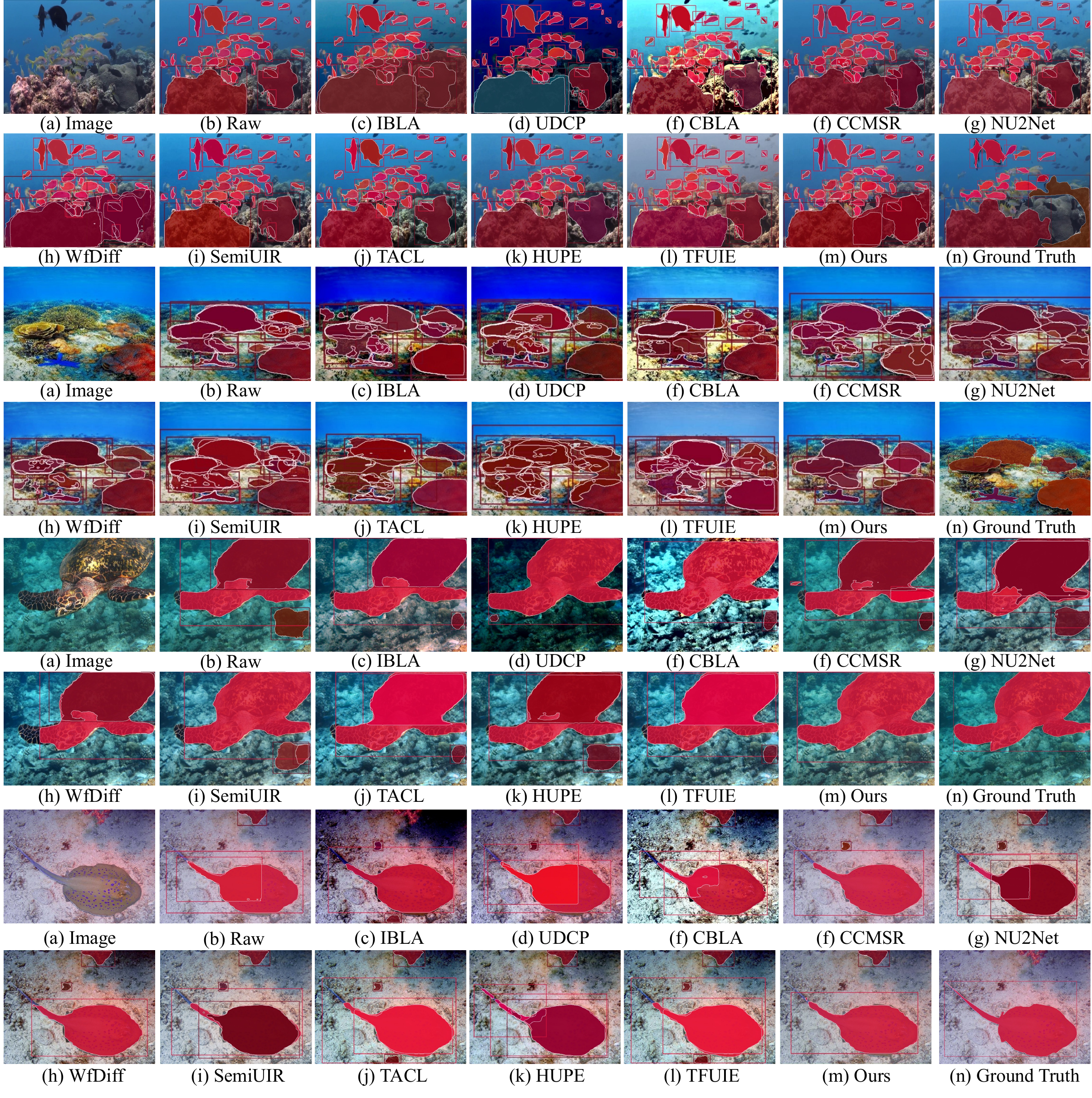}
  \caption{Visualization results of instance segmentation on UIIS dataset\cite{lianWaterMaskInstanceSegmentation2023}. The results are obtained by training the Mask R-CNN \cite{renFasterRCNNRealTime2016} network on the raw image and different enhanced images.}
  \label{Figure:intresult}
\end{figure*}

Table \ref{Table:instance} presents the quantitative results on the UIIS dataset. The proposed DTI-UIE provided +0.40 and +0.30 points of AP improvement over Mask R-CNN and Water Mask R-CNN, respectively. This fully demonstrates the effectiveness of our DTI-UIE in improving image feature representation, compared to both traditional and deep-learning-based UIE methods that fail on this task. The instance detection results are visualized in Figure \ref{Figure:intresult}.  From the visualization results, it can be seen that the proposed DTI-UIE enables the Mask R-CNN network to better distinguish the subtle relationship between the target and the background. Thus, the detection bounding box is more accurately selected and the segmentation mask is closer to the ground truth.

\begin{table*}[ht!]
\caption{Comparison for instance segmentation in terms of mAP50($\uparrow$), mAP75($\uparrow$) and mAP($\uparrow$) on UIIS\cite{lianWaterMaskInstanceSegmentation2023} dataset. The best and second result are marked in \textbf{bold} and \underline{underline}. }\label{Table:instance}
\centering
\setlength{\tabcolsep}{4pt} 
\scalebox{0.85}{
\begin{tabular}{l|cccccc|cccccc}
\toprule
\multirow{2}{*}{Method} & \multicolumn{6}{c|}{Mask R-CNN \cite{renFasterRCNNRealTime2016}}   & \multicolumn{6}{c}{Water Mask R-CNN \cite{lianWaterMaskInstanceSegmentation2023}} \\ 
\cmidrule{2-13}
& mAPs & mAPm  & mAPl & mAP50 & mAP75 & mAP   & mAPs & mAPm  & mAPl & mAP50 & mAP75 & mAP \\ 
\midrule
Raw      & \underline{7.50} & 17.70 & 29.70 & 40.10 & 20.50 & \underline{21.90} & \textbf{8.30} & 19.00 & \textbf{34.40} & \underline{40.00} & \underline{23.90} & \underline{23.30} \\
UDCP \cite{drewsUnderwaterDepthEstimation2016}& 7.70 & 18.60 & 28.20 & 40.10 & 19.80 & 20.30 & 6.70 & 17.10 & 29.60 & 36.50 & 19.50 & 20.00 \\
IBLA \cite{pengUnderwaterImageRestoration2017} & 7.20 & 18.60 & 28.50 & 38.70 & 20.60 & 20.60 & 6.30 & 17.20 & 29.60 & 35.90 & 20.70 & 20.10 \\
CBLA\cite{jhaCBLAColorBalancedLocally2024}     & 6.70 & 17.70 & 28.30 & 38.70 & 19.90 & 20.60 & 7.10 & 18.60 & 32.20 & 38.50 & 22.40 & 21.90 \\ 
NU2Net\cite{guoUnderwaterRankerLearn2022}   & \textbf{7.60} & \underline{19.00} & 29.60 & 39.50 & \textbf{22.00} & 21.60 & 7.70 & 18.90 & 32.60 & 39.20 & 23.70 & 22.60  \\ 
WfDiff\cite{zhaoWaveletbasedFourierInformation2024}   & 7.40 & 17.90 & 31.50 & 39.60 & \underline{21.90} & 21.70 & 7.60 & 18.90 & 29.90 & 37.60 & 21.80 & 21.50 \\ 
SemiUIR\cite{huangContrastiveSemisupervisedLearning2023}  & 7.20 & 17.90 & \underline{30.00} & \underline{40.20} & 21.30 & 21.20 & \underline{7.90} & \underline{19.40} & 32.80 & \textbf{40.10} & 22.60 & 22.90 \\ 
CCMSR\cite{qiDeepColorCorrectedMultiscale2024} & 7.30 & 18.50 & 27.70 & 37.30 & 18.70 & 20.30 & \underline{7.90} & 19.00 & 31.40 & 39.30 & 21.80 & 22.40 \\ \midrule
HUPE\cite{zhangHUPEHeuristicUnderwater2025}     & 7.30 & \textbf{19.20} & 29.30 & 39.20 & 21.70 & 21.30 & 7.40 & 19.00 & 31.60 & 38.30 & \underline{23.90} & 22.40 \\
TFUIE\cite{yuTaskFriendlyUnderwaterImage2024}    & 7.00 & 17.50 & 28.30 & 38.70 & 19.70 & 20.90 & 6.80 & \textbf{19.50} & 29.80 & 38.40 & 22.20 & 21.90 \\
TACL\cite{liuTwinAdversarialContrastive2022a}     & 6.70 & 16.20 & 25.20 & 33.20 & 17.40 & 17.40 & 7.00 & 17.10 & 26.50 & 32.20 & 18.40 & 18.50 \\ \midrule
DTI-UIE (Ours)     & \underline{7.50} & 18.30 & \textbf{32.20} & \textbf{40.40} & 20.60 & \textbf{22.30} & \underline{7.90} & 19.20 & \underline{33.80} & \textbf{40.10} & \textbf{24.30} & \textbf{23.60} \\ 
\bottomrule
\end{tabular}}
\end{table*}

\subsection{Comparison on Image Quality Metrics}

Three no-reference metrics for underwater image quality assessment were utilized for assessing the image quality of the proposed DTI-UIE and other UIE methods. The images for evaluation are obtained from two traditional UIE datasets for human visual perception, SUIM-E\cite{qiSGUIENetSemanticAttention2022} and UIEB\cite{liUnderwaterImageEnhancement2020}. As is shown in Table \ref{Table:iqa}, DTI-UIE is comparable to other UIE methods on UIQM \cite{zhangUnreasonableEffectivenessDeep2018}, UCIQE \cite{zhangUnreasonableEffectivenessDeep2018} and NUIQ \cite{jiangUnderwaterImageEnhancement2022} metrics. It should be noted that these indicators more reflect the underwater image quality under human visual perception and cannot be associated with downstream tasks.

\begin{table}[ht!]
\caption{Comparison for image quality assessment in terms of UIQM\cite{zhangUnreasonableEffectivenessDeep2018}($\uparrow$), UCIQE\cite{zhangUnreasonableEffectivenessDeep2018}($\uparrow$) and NUIQ\cite{jiangUnderwaterImageEnhancement2022}($\uparrow$) on SUIM-E\cite{qiSGUIENetSemanticAttention2022} and UIEB\cite{liUnderwaterImageEnhancement2020}. The best and second result are marked in \textbf{bold} and \underline{underline}. }\label{Table:iqa}
\centering
\scalebox{0.85}{
\begin{tabular}{l|ccc|ccc}
\toprule
\multirow{2}{*}{Method} & \multicolumn{3}{c|}{SUIM-E\cite{qiSGUIENetSemanticAttention2022}}  & \multicolumn{3}{c}{UIEB\cite{liUnderwaterImageEnhancement2020}} \\
\cmidrule{2-7} 
 & UIQM  & UCIQE  & NUIQ & UIQM  & UCIQE  & NUIQ \\ 
\midrule
UDCP\cite{drewsUnderwaterDepthEstimation2016} & 1.81 & 62.17 & 6.34 & 2.14 & 59.49 & 6.06 \\
IBLA\cite{pengUnderwaterImageRestoration2017} & 1.87 & 62.49 & 6.72 & 2.34 & 53.81 & 6.34 \\
CBLA\cite{jhaCBLAColorBalancedLocally2024}  & 2.24 & \textbf{67.44} & \textbf{6.75} & 2.32 & \textbf{68.72} & \textbf{7.45} \\ 
NU2Net\cite{guoUnderwaterRankerLearn2022}   & \textbf{3.00} & 58.41 & 5.74  & \textbf{3.34} & 58.07 & 5.99 \\ 
WfDiff\cite{zhaoWaveletbasedFourierInformation2024}   & 2.66 & 58.35 & \underline{6.45} & 3.25 & 53.26 & 6.57\\ 
SemiUIR\cite{huangContrastiveSemisupervisedLearning2023}  & 2.76 & 62.36 & 6.20 & 3.27 & 61.75 & \underline{6.72} \\ 
CCMSR\cite{qiDeepColorCorrectedMultiscale2024} & 2.72 & 57.48 & 6.26  & 3.33 & 59.97 & 5.75\\ 
HUPE\cite{zhangHUPEHeuristicUnderwater2025}     & 2.83 & \underline{63.41} & 5.15 & 3.21 & \underline{63.22} & 5.45 \\
TFUIE\cite{yuTaskFriendlyUnderwaterImage2024}    & 2.85 & 55.52 & 5.64 & 3.28 & 53.53 & 5.93 \\
TACL\cite{liuTwinAdversarialContrastive2022a}     & \underline{2.96} & 62.36 & 6.29 & 3.32 & 61.51 & 6.29 \\ \midrule
DTI-UIE     & 2.79 & 54.18 & 5.94 & \underline{3.33} & 54.92 & 5.98 \\ 
\bottomrule
\end{tabular}}
\end{table}

\subsection{Complexity Analysis}

We further analyze the computational complexity of the proposed DTI-UIE in comparison with other deep-learning-based UIE methods, in terms of the number of parameters, and  floating-point operations (FLOPs). As reported in Table \ref{Table:complexity}, the proposed method has the moderate complexity. 

\begin{table}[ht!]
\caption{Comparison for different deep-learning-based UIE methods in terms of model size and computational complexity. }\label{Table:complexity}
\centering
\scalebox{0.85}{
\begin{tabular}{l|cc}
\toprule
Method & Params (M) & FLOPs (G)   \\ 
\midrule
NU2Net\cite{guoUnderwaterRankerLearn2022}   & 3.1 & 10.4  \\ 
WfDiff\cite{zhaoWaveletbasedFourierInformation2024}   & 100.5 & 369.2   \\ 
SemiUIR  & 7.1 & 36.45  \\ 
CCMSR\cite{qiDeepColorCorrectedMultiscale2024} & 21.6 & 43.3   \\ 
HUPE\cite{zhangHUPEHeuristicUnderwater2025}     & 12.1 & 162.2   \\
TFUIE\cite{yuTaskFriendlyUnderwaterImage2024}    & 1.3 & 25.26   \\
TACL\cite{liuTwinAdversarialContrastive2022a}     & 11.3 & 199.8  \\ \midrule
DTI-UIE (Ours)     & 9.5 & 361.0   \\ 
\bottomrule
\end{tabular}}
\end{table}

\subsection{Ablation Study}
In the ablation studies, we analyze the effectiveness of the components of the proposed DTI-UIE network. The comparison is conducted using the UNet model for the semantic segmentation task on the SUIM dataset.

\textbf{Effectiveness of the two branch network.} Inspired by the advanced and primary visual processing mechanisms of human visual perception, FRB and DEB are designed to enhance target-relevant features and restore edge information, respectively. The impact of these two branches is shown in Table \ref{Table:ablationbranch}. The experimental results demonstrate that combining both branches improves the performance of the DTI-UIE on downstream tasks.

\begin{table}[]\centering
\caption{\textbf{Effectiveness of the FRB and DEB.} The comparison by the implementation of different network branches.}\label{Table:ablationbranch}
\scalebox{0.85}{
\begin{tabular}{l|ll}
\toprule
Methods                 & Dice  & mIoU    \\ \midrule
DTI-UIE \textit{without} FRB branch    & 72.89 & 67.01   \\ 
DTI-UIE \textit{without} DEB branch   & 72.52 & 66.41   \\ 
DTI-UIE (\textbf{Ours})         & \textbf{74.48} & \textbf{67.61}   \\
\bottomrule
\end{tabular}}
\end{table}

\textbf{Effectiveness of the target-relevant prior knowledge and TA-CTB block.} To enable the network to leverage task-relevant features, we introduce TA-CTB blocks to inject task-related priors into the network's encoder. To verify the effectiveness of this approach, we replace the TA-CTB blocks in decoder with CTB and conduct comparative experiments, as shown in Table \ref{Table:TACTB}. The experimental results demonstrate that incorporating target priors helps to produce images that are beneficial for downstream tasks, resulting in gains of +3.33 in Dice score and +0.48 in mIoU score.

\begin{table}[]\centering
\caption{\textbf{Effectiveness of the TA-CTB.} The comparison by including or excluding TA-CTB and target-relevant prior.} \label{Table:TACTB}
\scalebox{0.85}{
\begin{tabular}{l|ll}
\toprule
Methods                 & Dice  & mIoU    \\ \midrule
DTI-UIE \textit{without} TA-CTB    & 71.15 & 67.13   \\ 
DTI-UIE (\textbf{Ours})            & \textbf{74.48} & \textbf{67.61}   \\
\bottomrule
\end{tabular}}
\end{table}

\textbf{Effectiveness of the task-inspired perceptual loss.} 
To evaluate the role of the propose TDP loss, two complementary ablation settings are conducted. First, the DTI-UIE is trained with and without TDP loss while keeping all other configurations identical. As shown in Table \ref{Table:TDPloss}, incorporating TDP loss substantially improves downstream segmentation performance on the enhanced images, increasing mIoU by +2.49 and Dice by +2.77, which confirms that feature-level task-aware supervision is critical for producing task-friendly enhancement results. Second, whether updating TDP loss is sufficient is investigated  by training a Stage 1+2 variant that still uses TDP loss but computes with a pretrained and frozen task network. Compared with this variant, the full three-stage training yields consistently better downstream performance, indicating that updating TDP loss in Stage 3 provides additional gains and helps better align the enhancement outputs with the downstream task representation space. This advantage can be attributed to the following aspects: First, TDP loss optimizes the enhancement network in the representation space of a task model, so introducing and updating the task network can better adapt the loss to the enhanced domain, making the feature-space alignment more faithful and the supervision signal more effective. Second, the task network is jointly trained with enhanced images as well as mixed samples, which improves generalization and prevents the TDP loss from overfitting to artifacts introduced by enhancement. Overall, these results demonstrate that TDP loss provides effective task-aware guidance for producing task-friendly enhancements, and that updating the task network further strengthens this alignment and robustness.

Further, the impact of varying the granularity of image mixing on the performance of the TDP loss is examined. The default $8\times8$ block mixing proved the best balance between introducing diversity and maintaining structural coherence, leading to stable performance and effective task-oriented enhancements. Larger mixing blocks introduced boundary artifacts disrupting semantic consistency. The $1\times1$ caused excessive smoothing in high-frequency details, losing important discriminative features. Full-image mixing reduced local feature diversity, weakening task-specific performance. Overall, $8\times8$ block mixing method proved the most effective, minimizing artifacts and maintaining strong feature preservation.

\begin{table}[]\centering
\caption{\textbf{Effectiveness of the TDP loss.} The comparison by including or excluding TDP loss, as well as by not updating TDP loss.}\label{Table:TDPloss}
\scalebox{0.85}{
\begin{tabular}{l|ll}
\toprule
Methods                 & Dice  & mIoU    \\ \midrule
DTI-UIE \textit{without} TDP loss    & 71.71 & 65.12   \\ 
DTI-UIE \textit{with} no trainable TDP loss & 73.27 & 66.50 \\ 
DTI-UIE (\textbf{Ours})              & \textbf{74.48} & \textbf{67.61}   \\
\bottomrule
\end{tabular}}
\end{table}

\begin{table}[]\centering
\caption{\textbf{Effectiveness the mixing process in TDP loss.} The comparison by varying the granularity of image mixing on TDP loss.}\label{Table:TDPlossMix}
\scalebox{0.85}{
\begin{tabular}{l|ll}
\toprule
Methods                 & Dice  & mIoU    \\ \midrule
Mixing with $16\times16$ block & 72.96  & 65.94   \\ 
Mixing with $8\times8$ block (\textbf{Ours})& \textbf{74.48} & \textbf{67.61} \\ 
Mixing with $1\times1$ pixel   & 73.90 & 65.53   \\
Full image mixing & 73.50 & 65.32\\
No mixing in TDP loss & 71.34 & 64.82 \\
\bottomrule
\end{tabular}}
\end{table}

\begin{table}[]\centering
\caption{\textbf{Effectiveness of Task Network Number.} The comparison by merging or separating $\mathbf{Seg}_{pri}$ and $\mathbf{Seg}_{task}$ network.}\label{Table:Tasknetwork}
\scalebox{0.85}{
\begin{tabular}{l|ll}
\toprule
Methods                 & Dice  & mIoU    \\ \midrule
\textit{Merging} Task Networks   & 69.58 & 65.59   \\ 
\textit{Separating} Task Networks(\textbf{Ours}) & \textbf{74.48} & \textbf{67.61}   \\
\bottomrule
\end{tabular}}
\end{table}

\textbf{Effectiveness of the task networks.} 
To train the proposed DTI-UIE network, two task networks are designed to extract task-relevent priors and to compute the TDP loss, respectively. This decoupled design provides a stable source of task-related features from original image and simultaneously delivers a reliable task-driven supervision signal.  In contrast, when a single task network is used to generate both the priors and the TDP loss, the optimization becomes tightly coupled and can form a self-reinforcing feedback loop. Specifically, the prior distribution becomes non-stationary, which destabilizes training and weakens convergence. More importantly, the joint network can exhibit shortcut behavior. The enhancement network may reduce the loss by selectively suppressing or smoothing difficult regions instead of recovering discriminative structures, while the task network gradually adapts to these self-generated enhancement patterns. This leads to representation drift and increasingly impure priors that reflect the model's artifacts rather than task-relevant cues. The performance drop in Table \ref{Table:Tasknetwork} is consistent with the above mechanism. Separating the two task networks keeps the injected priors more stable and informative, prevents the task network used for supervision from over-fitting to the evolving outputs, resulting in more robust task-oriented enhancement.

Further, we attempted to change the structure of the task networks $\mathbf{Seg}_{pri}$ and $\mathbf{Seg}_{task}$ from UNet with VGG-16 baseline to ResNet50 and VGG-13. The results are shown in Table \ref{Table:Taskstructure}. The results show that the proposed network is robust to the structure of the task network, which fully demonstrates that the three-stage training framework can effectively concentrate on improving of  enhancement on downstream tasks.

\begin{table}[]\centering
\caption{\textbf{Effectiveness of Task Network Structure.} The comparison by the structure of $\mathbf{Seg}_{pri}$ and $\mathbf{Seg}_{task}$ network.}\label{Table:Taskstructure}
\scalebox{0.85}{
\begin{tabular}{l|ll}
\toprule
Methods & Dice  & mIoU    \\ \midrule
ResNet50 UNet  & \textbf{74.58} & 66.44   \\ 
VGG-13 UNet    & 71.64 & 66.90   \\ 
VGG-16 UNet (\textbf{Ours}) & 74.48 & \textbf{67.61}   \\
\bottomrule
\end{tabular}}
\end{table}

\begin{table}[]\centering
\caption{\textbf{Effectiveness of the TI-UIED.} The comparison by training on TI-UIED dataset or SUIM-E dataset.}\label{Table:TAUIED}
\scalebox{0.85}{
\begin{tabular}{l|ll}
\toprule
Methods                 & Dice  & mIoU    \\ \midrule
DTI-UIE \textit{trained on} SUIM-E    & 70.47& 65.18\\ 
DTI-UIE \textit{trained on} TI-UIED (\textbf{Ours}) & \textbf{74.48} & \textbf{67.61}   \\
WfDiff \textit{source pretrained} & 72.37 & 66.09 \\
WfDiff \textit{trained on} TI-UIED & 73.61 & 66.25 \\
CCMSR \textit{source pretrained} & 72.58 & 65.47 \\
CCMSR \textit{trained on} TI-UIED & 70.50 & 63.96 \\
\bottomrule
\end{tabular}}
\end{table}

\textbf{Effectiveness of the TI-UIED dataset.} To train the proposed DTI-UIE network, we created the TI-UIED dataset, which was developed using a perception and voting process based on downstream tasks. To evaluate the effectiveness of TI-UIED, we fine-tuned existing UIE models, such as WfDiff and CCMSR, on the dataset. We further train the DTI-UIE network on a human-oriented SUIM ground truth \cite{qiSGUIENetSemanticAttention2022}. The results, shown in Table \ref{Table:TAUIED}, show that while fine-tuning on TI-UIED led to some improvements, the performance gains were not consistent across all models. Specifically, CCMSR and WfDiff, which were originally designed for human visual perception, did not show significant improvements in downstream task performance after training on TI-UIED. This may be due to this model focuses on optimizing visual quality, rather than task-specific features. Besides, training DTI-UIE on SUIM-E significantly reduces the performance of downstream tasks, further supporting the idea that TI-UIED's dataset-level alignment enhances task-specific performance. The findings also reinforce that TI-UIED and DTI-UIE complement each other by preserving task-relevant features that are crucial for downstream task performance.

\section{Discussion and Limitation}\label{Section:6}
This study addresses a recurring mismatch in UIE tasks, which is the improvements in subjective visual quality do not equate to better performance for downstream perception tasks. In contrast to prior UIE methods that primarily optimize for human visual perception, the proposed DTI-UIE framework reframes enhancement as a task-oriented problem. Specifically, the task-inspired dataset construction strategy aligns supervision with downstream segmentation task by selecting enhancement outputs that consistently improve task performance across multiple models, thereby moving from visual perception-centered to task-centered. Besides, the DTI-UIE network decomposes the UIE problem into complementary pathways for global restoration for semantic and local reinforcement for details, matching to the requirements of tasks that depend on boundary fidelity and high-frequency structure. Task priors are incorporated through cross-branch attention fusion process, thereby emphasize features that are most informative for tasks. Finally, the staged training strategy with task-driven perceptual objective encourages feature-space alignment with task-relevant representations rather than generic perceptual similarity. Overall, the key contribution of DTI-UIE is not only performance gains on tasks, but a generalization perspective, which the UIE can be more effective when evaluation targets, learning signals and input supervision are explicitly calibrated to the intended downstream usage.

Despite these advantages, several limitations remain. First, the reference images in the task-inspired dataset are selected from a candidate pool produced by diverse UIE methods covering representative enhancement behaviors, and are chosen according to performance under an architecturally varied ensemble of semantic segmentation networks. We adopt semantic segmentation as the primary selection task because it is a representative dense prediction problem that simultaneously demands accurate global semantic understanding and faithful preservation of local spatial structures. This multi-model, segmentation-driven selection protocol reduces the risk that reference assignment is dominated by the inductive bias of any single enhancement algorithm or segmentation backbone, and instead favors enhancement outputs that yield consistent improvements across models. Nevertheless, because the supervision is tailored to segmentation, the resulting references may still exhibit a degree of task-specific bias. And in principle, alternative downstream objectives could shift the relative ranking of enhancement candidates for some images, particularly in borderline cases where multiple candidates achieve comparable scores. Encouragingly, our experiments show that the learned enhancement model transfers well to other downstream tasks, including object detection and instance segmentation, suggesting that segmentation-driven supervision captures task-relevant image properties that are not narrowly confined to segmentation alone. Even so, a promising direction for future work is to incorporate multiple downstream tasks more explicitly into both learning and dataset construction—for example, by jointly optimizing losses from segmentation, detection, and instance segmentation, and by selecting reference candidates through voting or aggregated criteria across tasks. Such multi-objective optimization and cross-task dataset construction could further improve robustness and better characterize practical transferability under diverse application requirements. In addition, we plan to apply the same task-oriented construction method to other underwater datasets with complementary distributions to expand the data base. This multi-dataset construction strategy will reduce dependence on biases in any single dataset and is expected to further enhance the model's robustness and practical transferability in heterogeneous underwater environments.

\section{Conclusion}\label{Section:7}

In this paper, we design a UIE framework that aims to provide efficient preprocessed images for downstream target recognition tasks, using human visual perception models as inspiration. Considering the primary and advanced visual mechanisms, we design a dual-branch enhancement network, using the FRB and the DEB to process target-related features and restore edge details, respectively. Since human vision can obtain prior knowledge from related tasks, we introduce TA-CTB and mix target-related prior knowledge into the network. The network uses a three-stage training framework, with a TDP loss. The network's training data, namely TI-UIED, is automatically constructed by the task networks with reference to a human evaluation database. Extensive experimental results demonstrate that the proposed DTI-UIE framework can be generally and effectively applied to underwater image recognition tasks.

\bibliographystyle{IEEEtran}
\bibliography{bibfile}

\end{document}